%% file: vlm.tex
\pdfoutput=1
\documentclass[UTF8]{article}
\PassOptionsToPackage{numbers, compress}{natbib}

\usepackage{makecell}
\usepackage[preprint]{neurips_2023}

\setcitestyle{numbers,square}
\usepackage{algorithm}
\usepackage{algorithmic}
\usepackage[utf8]{inputenc} % allow utf-8 input
\usepackage[T1]{fontenc}    % use 8-bit T1 fonts
\usepackage{hyperref}       % hyperlinks
\usepackage{url}            % simple URL typesetting
\usepackage{booktabs}       % professional-quality tables
\usepackage{amsfonts}       % blackboard math symbols
\usepackage{nicefrac}       % compact symbols for 1/2, etc.
\usepackage{microtype}      % microtypography
\usepackage{xcolor}         % colors
\usepackage{color}
\usepackage{enumitem}
\usepackage{graphicx}
\usepackage{inconsolata}
\usepackage{relsize}
\usepackage{amsfonts}
\usepackage{url}
\usepackage{graphicx}
\usepackage{subfigure}
\usepackage{caption}
\usepackage{hhline}
\usepackage{bm}
\usepackage{colortbl}
\usepackage{tabularx}
\usepackage{amsfonts}
\usepackage{amssymb}
\usepackage{amsthm,amsmath}
\usepackage{mathrsfs}
\usepackage{setspace}
\usepackage{CJKutf8}  %中文
\usepackage{subcaption}
\usepackage{multirow}
\usepackage{amssymb,amsthm,dsfont,mathrsfs}
\usepackage{amsmath}
\usepackage{hyperref}

\definecolor{cblue}{rgb}{0.21,0.49,0.74}
\definecolor{pos}{rgb}{0.57,0.73,0.93}
\definecolor{neg}{rgb}{0.93,0.74,0.78}

\hypersetup{breaklinks=true,colorlinks,citecolor=cblue}

\newcommand*{\img}[1]{%
    \raisebox{-0.1\baselineskip}{%
        \includegraphics[
        height=20pt,
        keepaspectratio,
        ]{#1}%
    }%
}

\title{\img{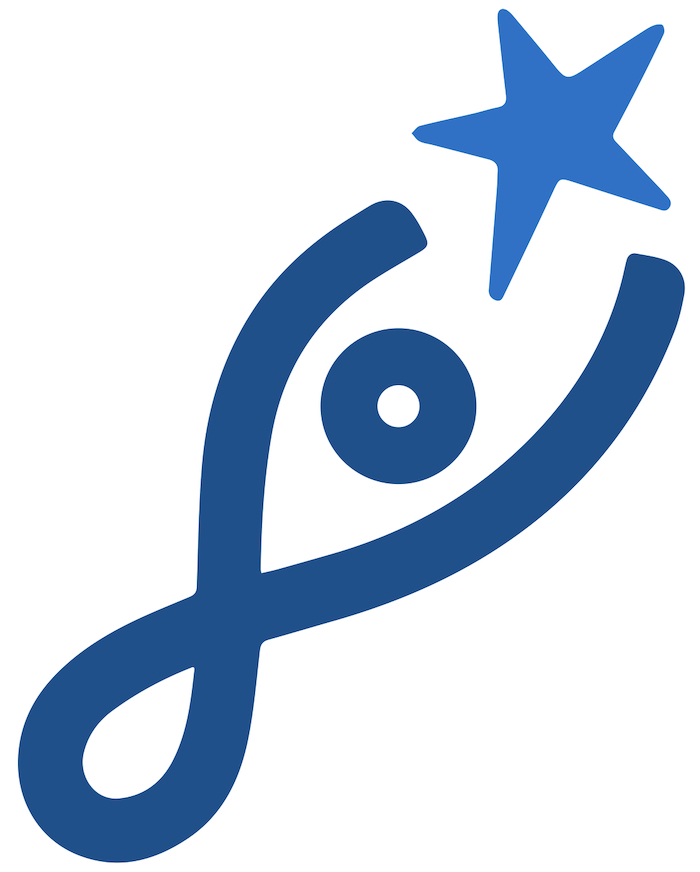}MouSi: Poly-Visual-Expert Vision-Language Models}

\author{%
  Xiaoran Fan\thanks{{ }\ Equal contributions.}\; , \; Tao Ji$^{*}$, \; Changhao Jiang$^{*}$, \; Shuo Li$^{*}$, \; Senjie Jin$^{*}$, \;  \\
  \vspace{2.5pt}
  \\
  \textbf{Sirui Song, \; Junke Wang, \; Boyang Hong, \; Lu Chen, \;} \\ 
  \textbf{Guodong Zheng, \; Ming Zhang, \; Caishuang Huang, \;} \\
  \textbf{Rui Zheng, \;  Zhiheng Xi, \; Yuhao Zhou, \; Shihan Dou, \; Junjie Ye, \; Hang Yan,  \;} \\
  \vspace{2pt}
  \\
  \textbf{Tao Gui$^{\dag}$, Qi Zhang\thanks{{ }{ }Correspondence to: \{tgui, qz\}@fudan.edu.cn},  Xipeng Qiu, Xuanjing Huang, Zuxuan Wu, Yu-Gang Jiang}\\
  \\
  \\
  \large Fudan NLP Lab \& Fudan Vision and Learning Lab \\ \\
}
\begin{document}

\maketitle
\makeatletter
\def\blfootnote{\xdef\@thefnmark{}\@footnotetext}
\makeatother
\newcommand{\todo}[1]{{\color{red}[{TODO:} #1]}}
\newcommand{\rebuttal}[1]{{\color{blue}{#1}}}

\begin{abstract}

Current large vision-language models (VLMs) often encounter challenges such as insufficient capabilities of a single visual component and excessively long visual tokens. These issues can limit the model's effectiveness in accurately interpreting complex visual information and over-lengthy contextual information. Addressing these challenges is crucial for enhancing the performance and applicability of VLMs. This paper proposes the use of ensemble experts technique to synergizes the capabilities of individual visual encoders, including those skilled in image-text matching, OCR, image segmentation, etc. This technique introduces a fusion network to unify the processing of outputs from different visual experts, while bridging the gap between image encoders and pre-trained LLMs. In addition, we explore different positional encoding schemes to alleviate the waste of positional encoding caused by lengthy image feature sequences, effectively addressing the issue of position overflow and length limitations. For instance, in our implementation, this technique significantly reduces the positional occupancy in models like SAM, from a substantial 4096 to a more efficient and manageable 64 or even down to 1. Experimental results demonstrate that VLMs with multiple experts exhibit consistently superior performance over isolated visual encoders and mark a significant performance boost as more experts are integrated. 

% We have open-sourced the training code used in this report. All of these resources can be found on our project website\footnote[1]{\ \ \url{https://github.com/FudanNLP/MOSS-MouSi}}.
% \end{abstract}

We have open-sourced the training code used in this report. All of these resources can be found on our project website\footnote[1]{\ \ \url{https://github.com/FudanNLPLAB/MouSi}}.
\end{abstract}

% our proposed 2D trainable image positional encoding alleviate the waste of positional encoding caused by lengthy image feature sequences, effectively addressing the issue of position overflow and length limitations. For instance, in our implementation, this technique significantly reduces the positional occupancy in models like SAM, from a substantial 4096 to a more efficient and manageable 64.

% Clip :Learning Transferable Visual Models From Natural Language Supervision 
% Dinov2: Learning robust visual features without supervision
% LayoutLMv3: Pre-training for Document AI with Unified Text and Image Masking
% Convnext: Convnext v2: Co-designing and scaling convnets with masked autoencoders.
% SAM: Segment Anything
% MAE: Masked Autoencoders Are Scalable Vision Learners

\input{outline/intro}
\input{outline/method}

\input{outline/experiment}

\input{outline/case}
\input{outline/related_work}

\section{Conclusion}

In this paper, we push the boundaries of vision-language models (VLMs) by proposing a novel polyvisual system that closely mirrors the complex and multi-dimensional nature of biological visual processing. Leveraging the unique attributes of diverse visual encoders, our system unifies their strengths to enrich the multimodal understanding of VLMs. Furthermore, we address the challenge of efficiently integrating visual information into language models by introducing techniques such as multi-patch-single-token projection and optimizing positional embeddings. This not only allows us to manage the overflow of vision tokens that typically burdens VLMs but also retains the models' semantic and spatial reasoning capabilities.
Through rigorous experiments across a suite of benchmarks, we demonstrate that our polyvisual approach significantly enhances the VLMs' performance, outpacing existing models in accuracy and depth of understanding. These results support our hypothesis that a well-integrated assembly of expert encoders can lead to a substantial improvement in handling complex multimodal inputs.

% \newpage
\bibliography{vlm}
\bibliographystyle{nips}

% \bibliographystyle{unsrtnat}
% \bibliography{main}

\newpage
\appendix
\include{outline/Appendix}

\end{document}

%% file: outline/intro.tex
\section{Introduction} \label{intro}

\input{figure/intro}

Current large vision-language models (VLMs) demonstrate significant potential in tasks requiring joint visual and linguistic perception, such as image captioning \cite{agrawal2019nocaps}, visual question answering \cite{antol2015vqa}, visual grounding \cite{yu2016modeling}, and autonomous agents \cite{durante2024agent,xi2023rise}. VLMs harness large language models (LLMs) as cognitive foundation models to empower various vision-related tasks, while \textbf{one vision component}, such as CLIP \cite{radford2021learning}, typically serves as auxiliary modules that provide additional visual perception \cite{liu2023llava}. However, the perception abilities of the individual vision models still lag behind, even in simple tasks like counting. \cite{yamada2022lemons,thrush2022winoground,yuksekgonul2022and}. This gap highlights a significant limitation in these models' capacity to process and understand visual information as effectively as they handle linguistic data. According to the operation of the vertebrate visual system, with each functional unit encoding different visual aspects in parallel, retinal ganglion cells transmit distinct features to the brain \cite{baden2016functional}. This biological mechanism suggests \textbf{a model structure where the varied visual information should be parallelly encoded by diverse perception channels. }

To this end, the community has verified that each model, with its unique approach to vision processing, contributes differently to understanding visual content \cite{chen2023vlp}. CLIP, with its contrastive learning approach, excels in aligning images with textual descriptions, providing a robust semantic understanding \cite{radford2021learning}. DINOv2, through its self-supervised learning paradigm at both the image level and patch level, offers significant advances in robust and stabilized feature extraction without relying on labeled data \cite{oquab2023dinov2}. LayoutLMv3's specialization in document AI tasks demonstrates the power of visual text processing \cite{huang2022layoutlmv3}. \cite{wang2023makes} empirically investigated different visual tokenizers pre-trained with dominant methods (i.e., DeiT \cite{touvron2021training}, CLIP, MAE \cite{he2021masked}, Dino \cite{caron2021emerging}), and observed that CLIP could capture more semantics, whereas the other models excelled at fine-grained perception. However, on the multimodal leaderboard organized by OpenCompass\footnote[2]{\ \ \url{https://rank.opencompass.org.cn/leaderboard-multimodal}}, the visual encoders of all open-source VLMs are based on the pre-trained CLIP encoder family. Many researchers have pointed out the shortcomings of the CLIP encoder, such as the inability to reliably capture even basic spatial factors of images \cite{kamath2023s}, suffering from object hallucination \cite{li2023evaluating}, and so on. In light of the distinct capabilities and limitations of these diverse vision models, a key question emerges: \textbf{How can we incorporate the strengths of multiple visual experts so that they work in synergy to improve overall performance?}

Drawing inspiration from biology, we take on the poly-visual-expert perspective and design a novel model, similar to how the vertebrate visual system operates. Consequently, in the process of developing VLMs with poly-visual experts, three problems are in major concern: (1) whether the poly-visual experts are effective; (2) how to better integrate multiple experts; and (3) how to avoid exceeding the LLM's maximum length with multiple visual experts? 

In order to verify whether multiple visual experts are effective for VLMs, we construct a candidate pool consisting of six well-known experts, including CLIP, DINOv2, LayoutLMv3, Convnext \cite{woo2023convnext}, SAM, and MAE. 
Using LLaVA-1.5 as the base setup, we explored single-expert, double-expert combinations, and triple-expert combinations in eleven benchmarks.
The results, as shown in Figure~\ref{fig:intro-exp-num}, indicate that as the number of visual experts increases, the VLMs acquire richer visual information (due to more visual channels), and the upper limit of the multimodal capability improves across the board.

In existing single visual channel VLMs, the methods for transmitting visual signals are either the MLP projection network \cite{liu2023improvedllava,wang2023cogvlm} or the Q-Former network \cite{bai2023qwen,dai2305instructblip}. 
To accommodate multi-channel signal transmission from multiple experts, we modified both methods for poly-expert fusion networks separately. 
The proposed method also compresses the local visual information by multi-patch-one-token for better transmission efficiency and reduces the quadratic computational cost of subsequent processing of VLMs.

In position-aware VLMs, vision tokens consume a staggering amount of positional embeddings.
Taking a single-turn multimodal dialogue in VQA as an example, with the MAE expert, the number of vision tokens (about 4096) is more than 500 times higher than the number of text tokens (about 8.7).
Inspired by the fact that visual experts already have positional encodings, we believe it is redundant to again assign a VLM position embedding to each visual token individually.
Therefore, we explore different positional encoding schemes to effectively address the issue of position encoding waste.
The results show that the two schemes: sharing one position for all patches and 2D positional encoding (rows plus columns) are able to reduce the position consumption (in the case of CLIP, the PE used drops from 576 to 24 or even 1), while the performance is still comparable.

Our contributions can be summarized as follows:
\begin{itemize}[noitemsep,leftmargin=*,topsep=0pt]
    \item We introduce a poly-visual-expert VLM that synergistically combines the strengths of various visual encoders to improve the overall capabilities of VLMs. 
    \item We tackle the challenge of vision token overflow in VLMs by proposing multi-patch-single-token projection and efficient positional encoding solutions. 
    \item By experimenting with different combinations of experts, our results demonstrate enhanced performance (+1.53 with fair comparison) in multimodal tasks. %超过在同样的训练数据下超过llava 1.5多少
\end{itemize}

%% file: figure/intro.tex
\begin{figure}[t!]
  \centering
  \includegraphics[width=\linewidth]{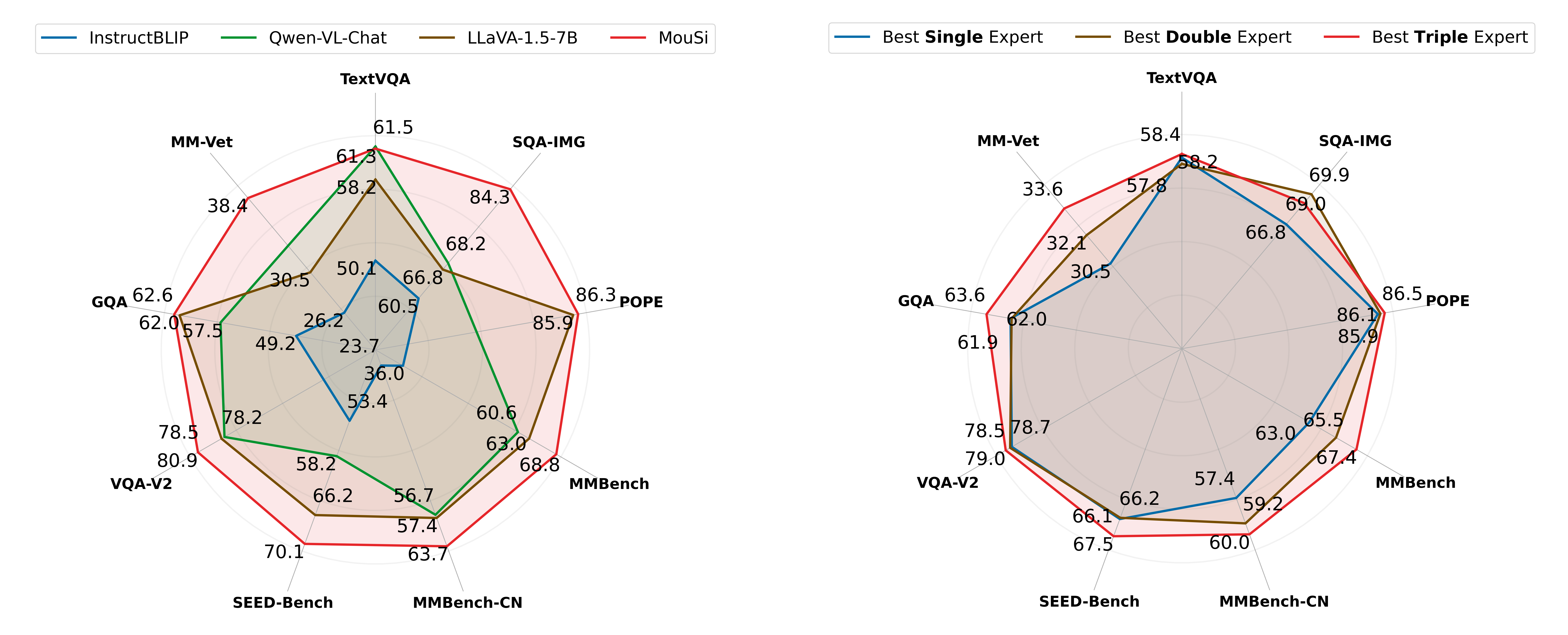}
  \caption{Left: Comparing InstructBLIP, Qwen-VL-Chat, and LLaVA-1.5-7B, poly-visual-expert \textbf{MouSi} achieves SoTA on a broad range of nine benchmarks. Right: Performances of the best models with different numbers of experts on nine benchmark datasets. Overall, triple experts are better than double experts, who in turn are better than a single expert.}
  \label{fig:intro-exp-num}
\end{figure}

%% file: outline/method.tex
\section{Architecture}
\label{sec:2}

\input{figure/overview}

\subsection{The Overview}

When a user uploads an image of wind pollination in a conical inflorescence and asks ``Which cones make pollen?'' the image is processed in sequence through the encodings of the CLIP expert, the SAM expert, and the LayoutLM expert, yielding three sets of visual representations. 
Subsequently, a poly-expert fusion network compresses the multi-channel visual information and aligns it multimodally to the vision input tokens for MouSi. 
The user's question is processed into text tokens by the LLMs' Embedding layer. 
Finally, MouSi generates the correct answer ``Male cones make pollen.'' by employing its VQA capabilities to understand the vision-language question, and its OCR capabilities to recognize the answer text from the image.

In order to accomplish the above task, we propose MouSi, which consists of three fundamental components: 
\begin{enumerate}[noitemsep,leftmargin=*,topsep=0pt]
 \item a multi-expert visual encoder, which combines the experts selected from a pool;
 \item a poly-expert fusion network, which is implemented as a simple projection fusion method or a Q-Former fusion method~\cite{li2023blip};
 \item a pre-trained open-source LLM (e.g., \emph{Vicuna v1.5}).
\end{enumerate}
Figure~\ref{fig:overview} shows an overview of the MouSi architecture. 
The core of a Vision-Language Model (VLM) is typically an LLM which is pre-trained on large-scale textual corpus. In order to perceive the visual signals, a vision encoder and vision-language connection layer are adopted to separately extract the visual features and align them to the semantic space of LLM. 

The VLM takes as input a sequence comprised of interleaved text and image segments, denoted as $X = (\dots, T_1, I_1, T_2, I_2, \dots)$, where text fragments $T$ are processed by the tokenizer and embedding layer of the LLM, and image segments $I$ are fed to the vision encoder. To ensure the universality and generalizability of the vision encoder, it is common practice to freeze its pre-trained parameters. In this paper, we rethink the design of the visual encoder in VLMs and aim to improve its capability by ensembled experts.

% The design and implementation details of the various components are as follows:

\subsection{Multi-Expert Vision Encoder}
% Clip :Learning Transferable Visual Models From Natural Language Supervision 
% DINOv2: Learning robust visual features without supervision
% LayoutLMv3: Pre-training for Document AI with Unified Text and Image Masking
% Convnext: Convnext v2: Co-designing and scaling convnets with masked autoencoders.
% SAM: Segment Anything
% MAE: Masked Autoencoders Are Scalable Vision Learners

After extensive investigation, we choose six vision encoders skilled in different domains, including CLIP~\cite{radford2021learning}, DINOv2~\cite{oquab2023dinov2}, LayoutLMv3~\cite{huang2022layoutlmv3}, Convnext~\cite{woo2023convnext}, SAM~\cite{kirillov2023segment}, and MAE \cite{he2021masked}. They differ significantly from each other in terms of input resolution, hidden size, model type, model size, pre-training tasks, and training methods, as shown in Table~\ref{tab:vision_experts}.

\input{table/vision_experts}
\paragraph{CLIP} learns the image-text alignment through contrastive learning. It is pre-trained on a large-scale dataset consisting of 400M noisy image-text pairs sourced from the internet. The vision encoder of CLIP is a Vision Transformer (ViT) with 300M parameters. The input resolution is fixed to 336$\times$336, and the feature dimension is 1024.\footnote[3]{\url{https://huggingface.co/openai/clip-vit-large-patch14-336}}
%An input image is preprocessed to a fixed resolution of 336$\times$336 and outputs 576 patch representation vectors of dimension 1024.

\paragraph{DINOv2} trains a student network to mimic the behavior of a more powerful teacher network, without the need for any training labels. Two objective functions are utilized for self-supervised pretraining: an image-level object that constrains the CLS tokens from the student network and teacher network, and a patch-level object that is applied to the extracted representations of masked input. The Dinov2 vision encoder is a Vision Transformer (ViT) with 1.1B parameters. The input image is preprocessed to 224$\times$224 resolution and the hidden dimension is 1536\footnote[4]{\url{https://huggingface.co/facebook/dinov2-giant}}.

% , which randomly masks some patches of the student network's input (but not the teacher network's), and then calculates the cross-entropy loss for the features of each masked patch.

\paragraph{LayoutLMv3} pre-trains multimodal Transformers for Document AI with unified text and image masking. 
% It is pre-trained with a word-patch alignment objective to learn cross-modal alignment by predicting whether the corresponding image patch of a text word is masked. 
The simple unified architecture and training objectives make LayoutLMv3 a general-purpose model for both text-centric and image-centric Document AI tasks. The LayoutLMv3 vision encoder is a ViT architecture with 368M parameters. The input image is first preprocessed to the resolution of 224$\times$224 and then encoded to 1024-dimension patch embeddings.\footnote[5]{\url{https://huggingface.co/microsoft/layoutlmv3-large}}

\paragraph{Convnext} is a purely convolutional network (ConvNet) that introduces a fully convolutional masked autoencoder framework (FCMAE) and a new global response normalization (GRN) layer to ConvNeXt. ConvNeXt underwent pretraining on the ImageNet-22K dataset, significantly enhancing the performance of the pure ConvNet across various recognition benchmarks. The ConvNeXt vision encoder we used has 200M parameters. The input resolution is 384$\times$384 and the feature dimension is 768.\footnote[6]{\url{https://huggingface.co/laion/CLIP-convnext_large_d_320.laion2B-s29B-b131K-ft-soup}}

\paragraph{SAM} is trained on a large-scale segmentation dataset, comprising 11 million images and over 1 billion masks, and achieves impressive zero-shot generalization. It is designed to efficiently predict object masks from images with different types of prompts, e.g., text or point. SAM also adopts ViT as a vision encoder with 637M parameters. The input resolution and hidden dimension are both larger, i.e., 1024$\times$1024 and 1280, respectively.\footnote[7]{\url{https://huggingface.co/facebook/sam-vit-huge}}
%By leveraging a powerful encoder and a lightweight decoder, SAM is capable of handling a wide range of segmentation tasks, making it an expert in the field of segmentation. 

%is a denoising self-supervised learning method that can An encoder maps the observed signals to latent representations, which are then used by a decoder to reconstruct the original signal.
\paragraph{MAE} aims to reconstruct the original image given only partial observations (25\% of the patches). The ViT-Huge encoder paired with MAE achieved a new record at the time on the ImageNet-1K dataset with an accuracy of 87.8\% and generalized very well. The MAE vision encoder has 630M parameters, while input resolution and hidden dimension are 1024$\times$1024 and 1280.\footnote[8]{\url{https://huggingface.co/facebook/vit-mae-huge}}

Given a image $I$ in the input sequence and a vision expert encoder $e_i(\cdot)$, we can obtain the representation vectors of $n$ image patches:
\begin{equation}
\bm{v}_{1}^{i}, \bm{v}_{2}^{i},\dots, \bm{v}_{n}^{i} = e_i(I).
\end{equation}
Assuming we have three experts ($e_i(\cdot)\in \mathbb{R}^{n_i\times d_i}$, $e_j(\cdot)\in \mathbb{R}^{n_j\times d_j}$, $e_k(\cdot)\in \mathbb{R}^{n_k\times d_k}$), the final sequence of image representations $V_{I}$ is a concatenation of the three output sequences.
\begin{align}
    V_I & = e_i(I) \oplus e_j(I) \oplus e_k(I) \nonumber\\
    & = [ \bm{v}_{1}^{i}, \dots, \bm{v}_{n_i}^{i}, \bm{v}_{1}^{j}, \dots, \bm{v}_{n_j}^{j}, \bm{v}_{1}^{k}, \dots, \bm{v}_{n_k}^{k} ]
\end{align}
It is worth noting that each expert outputs a different number ($n_i$ vs. $n_j$ vs. $n_k$) and dimension ($d_i$ vs. $d_j$ vs. $d_k$) of representations, and we will handle these differences in the poly-expert fusion network. 
In addition, the order of the experts could also have an impact on the results, which we specifically evaluate in the ablation experiments (Section~\ref{ssec:expert-order}).

% CLIP        & 336x336   & 1024 & ViT& large & ITC & weakly-supervised\\
% Dinov2      & 224x224   & 1536 & ViT& giant& DINO+iBOTlosses+SwAV& self-supervised\\
% LayoutLMv3  & 224x224   & 1024 & ViT& large& MLM+MIM+WPA & self-supervised\\
% Convnext    & 384x384   & 768  & CNN& large& Image classification & supervised\\
% SAM         & 1024x1024 & 1280 & ViT&huge & mask prediction & supervised\\
% MAE         & 224x320   & 1280 & ViT& huge& Image reconstruction& self-supervised\\

\input{figure/me-fusion}

\subsection{Poly-Expert Fusion Network}

Since the dimension and number of output sequences are often different for different visual experts, a fusion network needs to be designed to unify the processing.
Following LLaVA \cite{liu2023llava} and BLIP \cite{li2022blip}, we propose an MLP projection fusion network and a Q-Former fusion network, respectively.

% We add a visual expert module to each layer to enable deep visual-language feature alignment. Specifically, the visual expert module in each layer consists of a QKV matrix and an MLP in each layer. The shapes of the QKV matrix and MLP are identical to those in the pretrained language model and initialized from them. The motivation is that each attention head in the language model captures a certain aspect of semantic information, while a trainable visual expert can transform the image features to align with the different heads, therefore enabling deep fusion.

\paragraph{MLP projection} is a 2-layer ($d_{in} \to d_{hidden} \to d_{out}$) multilayer perceptron network.
To simplify the processing and to share the knowledge among multiple experts, we set the hidden dimension ($d_{hidden}$) and the output dimension ($d_{out}$) equal to the dimension ($d_{model}$) of the LLM, and the second layer network ($\operatorname{MLP}^{(2)}: d_{hidden} \to d_{out}$) parameters are shared among all experts.
Given a specific expert $e_i(\cdot)$, the first layer network is defined as $MLP^{(1)}_i: d_i \to d_{hidden}$.
\begin{align}
    % H_I & =  \operatorname{MLP}^{(1)}_i \left( e_i(I)\right) \oplus \operatorname{MLP}^{(1)}_j \left( e_j(I)\right) \oplus \operatorname{MLP}^{(1)}_k \left( e_k(I)\right) \nonumber \\
    % V_I & = \operatorname{MLP}^{(2)} (H_I)
    V_I & = \operatorname{MLP}^{(2)} \left( \operatorname{MLP}^{(1)}_i \left( e_i(I)\right) \oplus \operatorname{MLP}^{(1)}_j \left( e_j(I)\right) \oplus \operatorname{MLP}^{(1)}_k \left( e_k(I)\right) \right)
\end{align}
In practice, multiple experts output a large number of vision tokens, which not only increases the computational cost and memory usage of the VLM but also tends to exceed the maximum length limit during inference.
Therefore, we propose \textbf{multi-patches-one-token} projection to proportionally reduce the number of tokens output by each expert. 
Since image signals have local or sparse properties, it is reasonable to use one token to represent neighboring patches.
Take $m$-patch-one-token for example, we make the input dimension of the first layer of the network m times ($\operatorname{MLP}^{(1)}: d_{in}\times m \to d_{hidden}$), and its hidden layer output vectors $\bm{h}_1^i, \bm{h}_2^i, \dots$ are defined as follows:
\begin{align}
\bm{h}_1^i = \operatorname{MLP}^{(1)} \left(
\begin{bmatrix} 
\bm{v}_1^i  \\ 
\bm{v}_2^i \\ 
\vdots \\ 
\bm{v}_m^i 
\end{bmatrix} \right), \quad
\bm{h}_2^i = \operatorname{MLP}^{(1)} \left(
\begin{bmatrix} 
\bm{v}_{m+1}^i  \\ 
\bm{v}_{m+2}^i \\ 
\vdots \\ 
\bm{v}_{2m}^i 
\end{bmatrix} \right), \dots
\end{align}
where the $[\vdots]$ notation denotes concatenation over the vector dimension.
The final number of vision tokens is reduced to $\frac{1}{m}$ of the original.
In practice, $m$ is typically set from 2 to 8, which reduces cost while usually not losing performance on downstream tasks. 
If $m$ is set too large, the information of the image might be lost.

\paragraph{Q-Former network} is a trainable Querying Transformer module and proposed to bridge the gap between a frozen image encoder and a pre-trained LLM. 
It extracts a fixed number of output features from the vision encoder, independent of input image resolution.
We create a set number of learnable query embeddings as input to the Q-Former. 
The queries interact with each other through self-attention layers, and interact with frozen image
features $e_i(I)$ through cross-attention layers.
The output queries of the last layer are projected to the input layer of the LLM.
We use the pre-trained parameters in BLIP-2 as initialization to accelerate convergence and, similar to the second layer MLP network, share the parameters among all experts.
Since the dimension of query embeddings is equal to 768, we add an additional linear transformation ($W_i\in \mathbb{R}^{d_i\times 768}$) for each expert.
\begin{align}
    V_I & =\operatorname{Q-Former} \left( W_i \left( e_i(I)\right) \oplus W_j \left( e_j(I)\right) \oplus W_k \left( e_k(I)\right) \right)
\end{align}
The ablation study in Section~\ref{ssec:fusion-method} shows that the MLP fusion network fuses better than the Q-Former despite having fewer parameters and not being pre-trained.

\input{figure/PE}

\subsection{Different Positional Encoding Schemes}
\label{ssec:2DPE}
% [TODO: move]

Although the $m$-patch-one-token operation or defining a small number of queries in the Q-Former has been able to reduce the proportion of vision tokens, the occupation of position embeddings by vision tokens should not be underestimated during inference. 
Inspired by the fact that visual experts already have positional encodings  (e.g., 2D position encoding in ViT \cite{wang2019translating}), we believe it is redundant to again assign a VLM position embedding to each visual token individually. 
As shown in Figure~\ref{fig:pe}, this report explores four positional encoding schemes for improving the assignment of position embeddings (PEs):
\begin{enumerate}[noitemsep,leftmargin=*,topsep=0pt]
    \item a separate position vector for each patch (\emph{original});
    \item all vision tokens of an image share a PE (\emph{share-all});
    \item one PE shared by the same row of vision tokens (\emph{share-by-row});
    \item one PE shared by the same row of vision tokens, plus a set of learnable columns PEs (\emph{share-by-row\&col}).
\end{enumerate}
Among the four methods, \emph{share-all} can reduce the original $O(N^2)$ PE cost to $O(1)$, while the \emph{share-by-row} and \emph{share-by-row\&col} can reduce the PE cost to $O(N)$. 
All of them can significantly alleviate the out-of-maximum-length problem, but the question is \textbf{how much do they affect the performance of VLM?} 
We report ablation results in Section~\ref{ssec:PE}.

%% file: figure/overview.tex
\begin{figure*}[t!]
  \centering
  \includegraphics[width=\textwidth]{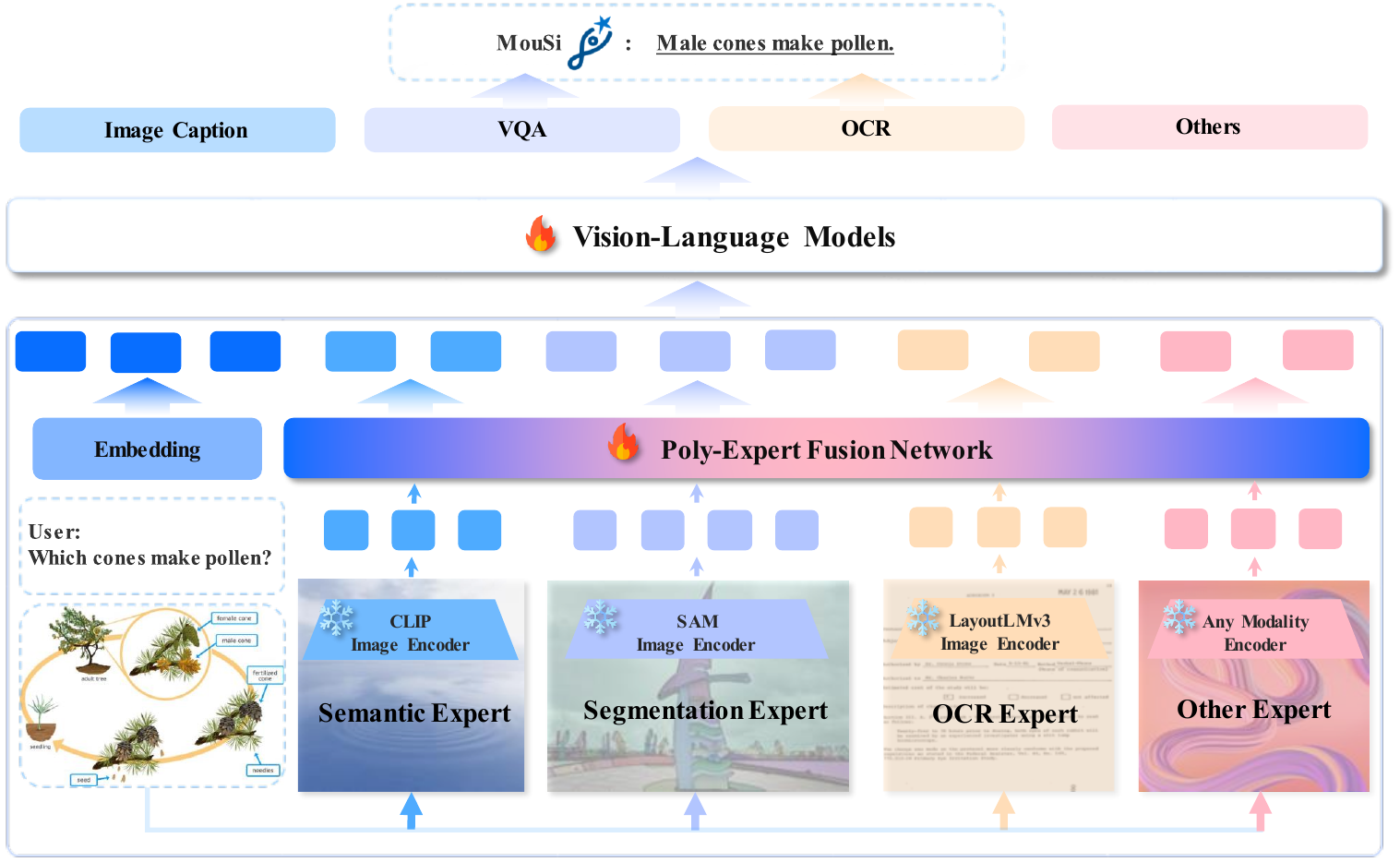}
  \caption{An overview of the MouSi model structure. 
  The poly-vision-expert MouSi model supports the integration of visual experts with various types and capabilities.
  % \url{https://1drv.ms/p/s!Ai1M4RXE1UjiikSPTAunEy_BPMAN}.
  }
  \label{fig:overview}
\end{figure*}

%% file: table/vision_experts.tex
\begin{table*}[t!]
\centering

\begin{tabular}{lrrrrclr}
\toprule %hidden_size, model type, model size, pre-training tasks, and training methods几个维度，用一段英文介绍MAE \cite{he2021masked} 图像编码器
% \multirow{2}{*}{\textbf{Expert}} & \multirow{2}{*}{\textbf{Res.}} & \multirow{2}{*}{\textbf{Param.}} & \textbf{Hidden} & \multirow{2}{*}{\textbf{\#Patch}} &  \textbf{Model} & \textbf{Pre-training}   \\
%     & & & \textbf{Size} & & \textbf{Type} &  \textbf{Tasks}  \\

\multirow{2}{*}{\textbf{Expert}} & \multirow{2}{*}{\textbf{Res.}} & \multirow{2}{*}{\textbf{Param.}} & \multirow{2}{*}{\textbf{d\_hid}} & \multirow{2}{*}{\textbf{\#Patch}} &  \multirow{2}{*}{\textbf{Type}}  & \multicolumn{2}{c}{\textbf{Pre-training}} \\
\cline{7-8}
                                 &                                &                                  &     &                                   &   &     \textbf{Tasks}    & \textbf{Images}   \\
% \multirow{2}{*}{\textbf{Expert}} & \multirow{2}{*}{\textbf{Res.}} & \multirow{2}{*}{\textbf{Param.}} & \textbf{Hidden} & \multirow{2}{*}{\textbf{\#Patch}} &  \textbf{Model} & \textbf{Pre-training} & \multirow{2}{*}{\textbf{Pretrain Images}} \\
%     & & & \textbf{Size} & & \textbf{Type} &  \textbf{Tasks} & \\
\midrule
CLIP       & 336  & 300M & 1024 & 576  & ViT & Image-Text Matching & 400M \\
DINOv2       & 224   & 1.1B & 1536 & 256  & ViT & DINO+iBOT+SwAV & 142M\\
LayoutLMv3  & 224   & 368M & 1024 & 196  & ViT & Document OCR & 11M \\
ConvNeXt    & 384   & 200M & 768  & 1024 & CNN & Image Classification & 2B \\
SAM         & 1024 & 637M & 1280 & 4096 & ViT &  Image Segmentation & 11M \\
MAE         & 224   & 630M & 1280 & 256  & ViT & Patch-level Denoising & 1.3M\\
\bottomrule
\end{tabular}
\caption{Comparison of six pre-trained visual experts. \textbf{Res.} indicates image resolution, \textbf{d\_hid} indicates hidden dimension and \textbf{Param.} indicates the number of parameters.}
\label{tab:vision_experts}
\end{table*}

%% file: figure/me-fusion.tex
\begin{figure}[t!]
  \centering
  \includegraphics[width=0.6\linewidth]{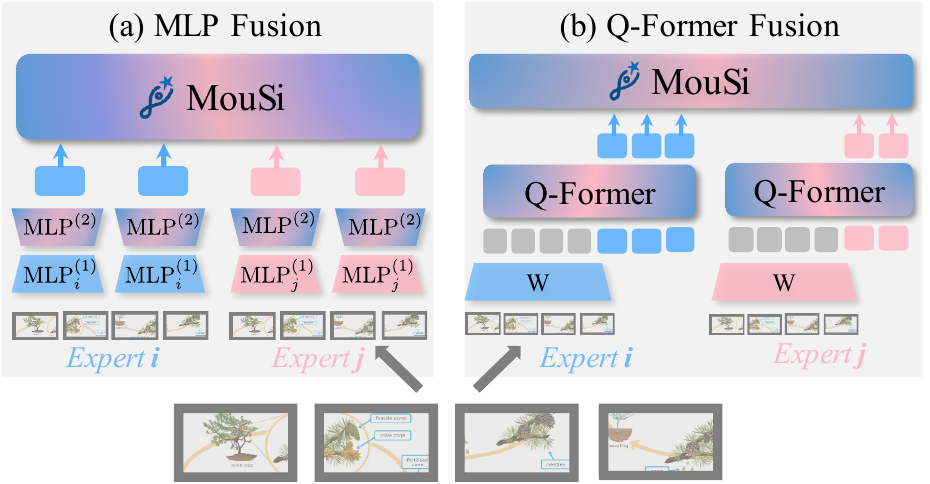}
  \caption{Examples of two types of multi-expert fusion networks. We show how the MLP method compresses visual information with ``2-patches-1-token'', and how the Q-Former method compresses information with 3 trainable queries. The modules with color gradients represent the sharing of parameters among multiple experts to transfer knowledge.
  % \url{https://1drv.ms/p/s!Ai1M4RXE1UjiikuNqzzXMwo4rsoF}
  }
  \label{fig:me-fusion}
\end{figure}

%% file: figure/PE.tex
\begin{figure*}[t!]
  \centering
  \includegraphics[width=\textwidth]{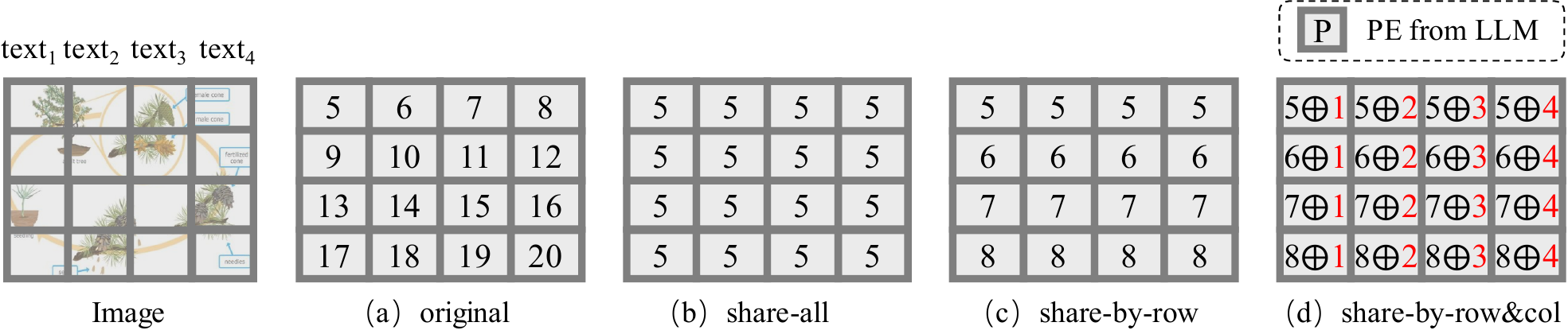}
  \caption{
  Diagram of the four positional encoding schemes. The $\oplus$ operator indicates that the row position embedding and column position embedding are summed.
  }
  \label{fig:pe}
\end{figure*}

%% file: outline/experiment.tex
\section{Experiments}
\label{sec:exp}
% 标题
\subsection{Main Results}
The main focus of our experiments is to conduct explorations of single-expert, double-expert, and triple-expert ensembles.
Following LLaVA-1.5 \cite{liu2023improvedllava}, our training pipeline consists of two phases. 
In phase 1, or the pre-training phase, we freeze the text-only LLM and the multi-expert encoder, and train the poly-visual fusion network from scratch to align the representation space of both. 
After training on a large-scale weakly-supervised (with noise) dataset, the text-only LLM is already capable of multimodal input and comprehension.
In phase 2, or the fine-tuning phase, we unfreeze the LLM and further train it together with the poly-visual fusion network on diverse and high-quality supervised fine-tuning (SFT) datasets.  
The construct of the datasets and the training configuration for both stages are detailed as follows.
\paragraph{Datasets.}
During the pre-training phase, we utilized the LCS-558K dataset, which comprises $\sim$558K image-text pairs from the LAION-CC-SBU collection, annotated with BLIP-generated captions.
During the fine-tuning phase, we mixed 10 diverse and high-quality SFT datasets containing VQA, OCR, region-level VQA, visual conversation, and language conversation data. 
To reduce training costs and enhance efficiency, we adopted the same preprocessing strategy as LLaVA-1.5, ultimately obtaining $\sim$665K SFT samples.
All data splits are concatenated together and sampled with the same probability. 
We selected 9 of the 12 evaluation benchmarks for LLaVA-1.5 (excluding LLaVA-Bench that rely on unstable GPT4 responses, as well as VisWiz \cite{gurari2018vizwiz} and MME \cite{fu2023comprehensive} for the website crashed), including VQA$^{\text{v2}}$ \cite{goyal2017making}; GQA \cite{hudson2019gqa}; SQA$^\text{I}$ : ScienceQA-IMG \cite{lu2022learn}; VQA$^\text{T}$: TextVQA \cite{singh2019towards}; POPE \cite{li2023evaluating}; MMB \& MMB$^{\text{CN}}$: MMBench \& MMBench-Chinese \textit{dev} results \cite{liu2023mmbench}; SEED$^\text{I}$ : SEED-Bench-IMG \cite{li2023seed}; MM-Vet \cite{yu2023mm}.
Detailed statistical information can be found in Appendix~\ref{app:data}.

\paragraph{Hyperparameters.} 
For main results, we keep all training hyperparameters roughly the same as the LLaVA series \cite{liu2023llava,liu2023improvedllava}.
% During the pre-training phase, 
% During the fine-tuning phase,
We present a detailed description of the hyperparameters in Appendix~\ref{app:hyper}.
For the MLP fusion network, we set $m$ in $m$-patches-one-token from 1 to 16 to avoid exceeding the maximum length for training and inference.
For the Q-Former fusion network, we set the number of queries per expert to match the number of outputs from the MLP fusion network. The parameters of the Q-Former fusion network are initialized using the pre-training parameters of BLIP-2 \cite{li2023blip}.

\subsubsection{Single Vision Expert}
\label{sec:exp-single}

\input{table/single_expert}

Table~\ref{tab:single-exp} compares the performance of all six VLMs with a single vision expert. 
The CLIP expert achieves the best performance in \textbf{all} 9 benchmarks, fully explaining why it has become the dominant choice of vision encoder for VLMs. 
Comparing the different attributes of the experts, CLIP ranked 5th in terms of the number of parameters, 3rd in terms of resolution, and 2nd above the size of the pre-training data, none of which had an absolute lead. 
Therefore, we guess that its main advantage lies in its image-text matching pre-training task, which has multimodal alignment capability in advance.
Overall, the performance ranking of the six experts is roughly CLIP$>$ConvNeXt$>$DINOv2$>$SAM$>$MAE$>$LayoutLMv3.
% Where ConvNeXt on VisWiz and MAE on MM-Vet surpass the CLIP expert, revealing the potential of multi-expert fusion.
In addition, LayoutLMv3 is an undisputed expert in OCR and SAM in image segmentation but performs poorly as a single visual encoder in VLM. 
A natural question is \emph{whether multi-expert fusion can activate their capabilities in their specialized fields?}

\subsubsection{Double Vision Experts}
\label{sec:exp-double}

\input{table/two_experts}

The current mainstream open-source VLMs have only one visual encoder, i.e., a single visual channel. However, multimodal tasks are diverse, and different tasks require different visual signals.
In this subsection, we investigate whether dual-channel, i.e., double visual experts can outperform single experts on various tasks.
We combine the strongest CLIP expert with other experts to construct a total of four double-expert combinations. 

Table~\ref{tab:double-exp} shows the performance of the double-expert vision encoder on the nine benchmarks, and relative to each single expert belong them (a positive number indicates that the double expert outperforms the single expert).
The results show that the ``DINOv2+CLIP'' experts, ``LayoutLMv3+CLIP'' experts, and ``ConvNeXt+CLIP experts'' three double-expert encoders outperform the arbitrary single encoder in almost all cases (23/27).
The results indicate that two visual channels do outperform a single visual channel in terms of multimodal capabilities, demonstrating that multi-expert collaboration is feasible.
For the ``SAM+CLIP'' combination, the results are surprising, with the dual expert outperforming the single expert in only 2/9 benchmarks, and lagging behind the single expert (specifically CLIP) in the remaining 7 benchmarks. The main reason might be that SAM encodes much more signals than CLIP (4096 patches vs. 576 patches), and fusion networks require a large information compression ratio. The most efficient CLIP channel is also compressed at this point, leading to performance decreases. 
% It is necessary to develop a more efficient information transfer network for visual experts with large amounts of information such as SAM.
There is a need to develop a more efficient visual information transfer network for experts with massive patches such as SAM.

Comparing the performance between double-expert methods, we found that the best double-expert is DINOv2+CLIP, rather than the ensemble of the best single expert and the second-best single expert, ConvNeXt+CLIP.
It indicates that superior performance as a single expert does not necessarily imply optimality when ensembled.
Since ConvNeXt and CLIP have considerable overlap in their training methods and training corpora, leading to the extraction of similar visual information, whereas the self-supervised DINOv2 and the weakly-supervised CLIP complement each other, resulting in a more effective ensemble.
Furthermore, it is worth mentioning that LayoutLMv3, which performed the worst as a single expert, shows significant improvement when collaborating with  CLIP, performing the best on four benchmarks and ranking overall just behind DINOv2+CLIP. 
Even SAM, whose information was compressed, achieved the highest performance on the ScienceQA-IMG benchmark. Therefore, we can conclude that when paired with the versatile visual expert CLIP, other experts can focus on capturing supplemental visual information to further enhance performance.

\subsubsection{Triple Vision Experts}
\input{table/three_experts}

Based on the double-expert encoder, we further construct the three-expert combinations.
As shown in Table~\ref{tab:triple-exp}, the three-expert approach wins in 4/6 cases in comparison with the two-expert at the data size of LLaVA-1.5.
The best-performing three-expert is LayoutLMv3+DINOv2+CLIP, followed by ConvNeXt+LayoutLMv3+CLIP, and finally ConvNeXt+DINOv2+CLIP.
Among them, model LayoutLMv3+DINOv2+CLIP has the largest number of parameters, reaching 8.8 billion.
We suspect that the main reason limiting the performance of the triple-expert methods is the insufficient amount of data.
We train the MouSi on larger (1647K) augmented data and observe more significant performance gains in Section~\ref{ssec:data-enhance}.

\subsection{Ablation Study}

\subsubsection{Effect of Fusion Methods}
\label{ssec:fusion-method}
\input{table/fusion_methods}

The MLP projection and Q-Former network are two mainstream methods for connecting vision and language. 
\emph{Which of them can more effectively convey visual signals} is a key issue, especially in the context of multi-expert fusion. 
Table~\ref{tab:fusion-exp} presents the performance of using MLP and Q-Former respectively on three double-expert combinations, including ``DINOv2 \& CLIP'' and ``ConvNeXt \& CLIP''. 
The results demonstrate that MLP significantly outperforms Q-Former in \textbf{all} cases, despite having fewer parameters and not utilizing pre-trained parameters like Q-Former, being instead directly initialized randomly.
It suggests that we should prefer a straightforward connection in the LLaVA with poly-visual experts setup.

\subsubsection{Effect of Expert Order}
\label{ssec:expert-order}
\input{table/experts_order}

Due to the autoregressive and position-aware characteristics of LLMs, even if the visual experts are exactly the same, a different order alone could affect the final output.
Table~\ref{tab:order-exp} presents the effect of swapping the order between double experts.
The swap results of groups ``DINOv2 \& CLIP'' and ``ConvNeXt \& CLIP'' indicate that order can cause some fluctuations in performance, with gains (7 of 22) on some benchmarks and losses (15 of 22) on others.
In general, placing CLIP later brings about better overall performance.
Because CLIP is the most effective single expert and the expert placed later is closer to the generation, we speculate that the latter-positioned expert has a slightly greater effect on the response.
This phenomenon is also consistent with the characteristics of binocular vision organisms, such as humans, where one eye is the dominant eye and the other is the non-dominant eye. The brain typically favors the input from the dominant eye when processing visual information \cite{miller1989ocular}.

\subsubsection{Effect of Different Positional Encoding Schemes}
\label{ssec:PE}
\input{table/PE-table}

This subsection compares the four positional encoding schemes of VLMs introduced in Section~\ref{ssec:2DPE}.
Table~\ref{tab:PE-exp} shows the results of the four approaches, where share-all not only saves the most PE but also improves the average performance by 0.8 on top of CLIP. The 2D positional coding (share-by-row\&col) also improves the average performance by 0.6.
However, share-by-row impairs the performance of the model, probably because row sharing corrupts the position information of the visual coder itself.
The experimental results validate our conjecture that it is redundant to re-assign LLM positional encoding to each vision token that already has positional information.

\subsection{Analysis}

\input{table/attention}

Among multiple visual encoders, one question worthy of analysis is the contribution of different experts to the model's output. 
Attention mechanisms are commonly used interpretive tools in Transformer networks. 
Here, we take a three-expert encoder as an example and analyze the average contribution of each expert across two multilingual benchmarks, MMB-English and MMB-Chinese. The contribution of one sample is the output token's average attention to each expert's representation.
Averaging over the entire dataset yields the overall average contribution.

Table~\ref{tab:atten-exp} shows the individual contributions of the text prompt, LayoutLMv3, DINOv2, and CLIP to the output. The results indicate that the contribution of the text prompt to the answer is significantly higher than that of the visual experts. This is as expected. Firstly, the text prompt defines the format of the VLM's response, necessitating attention to the prompt during output, and secondly, the text has a higher information density than images, hence the average attention is usually higher for text.
Comparing the three visual experts, we find that their contributions in descending order are CLIP, DINOv2, and LayoutLMv3. CLIP still demonstrates the characteristics of being the dominant eye or the primary visual channel.
DINOv2's contribution is approximately 20\% of CLIP's, while LayoutLM's contribution is minimal, at only 1\% of CLIP's.

\input{figure/qa_case}
A natural question that follows is, given the existence of visual channels with very low contributions, is there a necessity for them to be part of the model?
Figure~\ref{fig:qa_case} shows our perturbation experiments on the triple-expert LayoutLMv3+DINOv2+CLIP model. The output tokens of the corresponding expert are fully masked when generating answers, thus exploring the effect of the current expert on the output.
In Case 1, the user asks MouSi a simple question: ``Where is the dog in the picture?''. 
No matter which visual expert's output signal is masked, the remaining two visual channels are sufficient to correctly answer the location question ``on top of''. More details are provided when CLIP experts are present, such as outputting ``wooden table'' instead of just ``table''.
In Case 2, the user asks MouSi ``How many dogs are there in the picture? What colors are they?'' 
The perturbation results show that only three experts working together can answer the question correctly. The absence of any one expert results in an incorrect answer, which demonstrates the difference in the information captured by the multiple visual channels of the poly-visual-expert VLM. 
Some multimodal tasks rely on the synergy of multiple channels, which a single channel (i.e., a single expert VLM) does not have.

\subsection{Data Enhancement}
\label{ssec:data-enhance}
\input{table/extend-data}

After comprehensively exploring the architecture and effectiveness of the poly-visual-expert VLM, we further augmented the data from LLaVA-1.5 to explore the upper limits of the performance of the poly-visual-expert VLM.

\paragraph{Setting}  
During the pre-training phase, we used 1.2 million pre-training data to replace the original 558K data in LLaVA-1.5.
Where 100K data were generated by GPT4v, and the remaining data were produced by a supervised trained image captioner, which included the 558K images but with higher quality captions.
During the SFT phase, we expanded the 665K SFT data to 1647K. Detailed statistical information can be found in Appendix~\ref{app:data}.
For data enhancement results, we keep all training hyperparameters roughly the same as the main results.
Besides the number of iterations varies with the increase of data size.

Table~\ref{tab:enhance-exp} reports the results for LLaVA-1.5 (i.e., single CLIP expert), LLaVA-1.5 after data enhancement, and MouSi (with triple-expert ``LayoutLM+ConvNeXt+CLIP'') after data enhancement on nine benchmarks. 
The results show that with data enhancement, the poly-visual expert VLM can further improve performance (7/9) compared with single-expert VLM.
The average performance improved by 1.0, yet the number of parameters increased by only 300M.
Comparing the effects of data augmentation, we observe that the single-expert approach improved by 4.4, and the triple-expert method improved by 3.9. 
This confirms that the potential of poly-visual-expert VLMs has not yet been fully tapped and that more data can significantly enhance the capabilities of VLMs.
Finally, compared to mainstream VLMs, MouSi performs the best in 8 out of 9 benchmarks while exhibiting the second-best performance in the remaining one, demonstrating strong multimodal assistant capabilities.

%% file: table/single_expert.tex
\begin{table*}[h]
\fontsize{15}{20}\selectfont
\setlength{\tabcolsep}{3pt}
\centering

\resizebox{\textwidth}{!}{
\begin{tabular}{lrccccccccccccc}
\toprule

Model & Param.  & VQA$^{\text{v2}}$ & GQA  & SQA$^\text{I}$ & VQA$^\text{T}$ & POPE & MMB & MMB$^{\text{CN}}$ & SEED$^\text{I}$  & MM-Vet &Avg. \\
\midrule 
\rowcolor{gray!10} \multicolumn{12}{c}{\textit{\textbf{Single Expert}}} \\
% \multicolumn{19}{c}{\textit{Full Attention}} \\
% \textit{Full Attention}\\
CLIP       & 7.3B & \bf 78.5 & \bf 62.0  & \bf 66.8 & \bf 58.2 & \bf 85.9  & \bf 63.0 & \bf 57.4 & \bf 66.2  & \bf 30.5 &\bf63.2  \\
DINOv2     & 8.1B & 74.9 & \underline{61.7}  & \underline{66.1} & 46.2 & 84.6 & 57.9 & 48.7 & \underline{63.4}   & 23.4 &58.5 \\
LayoutLMv3 & 7.4B & 44.9 & 40.0 & 62.8 & 43.6 & 59.1 & 29.0 & 19.8 & 34.8   & 11.8  &38.4\\
ConvNeXt   & 7.2B & \underline{75.1} & 60.5 & 65.0 & \underline{56.3} & \underline{85.6} & \underline{63.3} & \underline{55.0} & 61.5 & \underline{26.0} &\underline{60.9} \\
SAM        & 7.6B & 64.7 & 55.8 & 63.9 & 44.1 & 82.0 & 43.7 & 33.9 & 51.9   & 17.7 &50.9 \\
MAE        & 7.6B & 62.0 & 53.2 & 63.3 & 44.5 & 79.7 & 41.6 & 33.0 & 49.4 &  16.5  &49.2\\
\bottomrule
\end{tabular}
}
\caption{{\bf Comparison of six vision experts on 9 benchmarks.} Param indicates the number of parameters.}
\label{tab:single-exp}
% \vspace{-15pt}
\end{table*}

%% file: table/two_experts.tex
% \begin{figure*}[t!]
%   \centering
%   \includegraphics[width=\textwidth]{figure/two-experts-crop.pdf}
%   \caption{
%   \url{https://1drv.ms/p/s!Ai1M4RXE1UjiikSPTAunEy_BPMAN}.
%   }
%   \label{fig:two-experts}
% \end{figure*}
\begin{table*}[ht]
% \fontsize{15}{20}\selectfont
\setlength{\tabcolsep}{3pt}
\centering
\includegraphics[width=\textwidth]{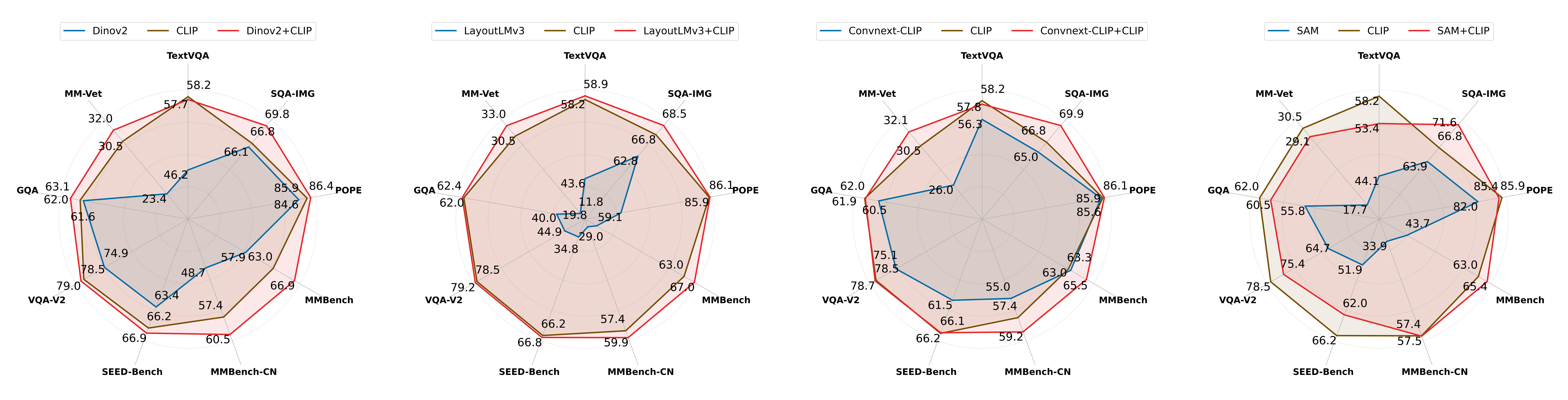}
\resizebox{\textwidth}{!}{
\begin{tabular}{lrrrrrrrrrrrrrr}
\toprule
Model & Param.  & VQA$^{\text{v2}}$ & GQA & SQA$^\text{I}$ & VQA$^\text{T}$ & POPE & MMB & MMB$^{\text{CN}}$ & SEED$^\text{I}$  & MM-Vet & Avg. \\
\midrule 
\rowcolor{gray!10} \multicolumn{12}{c}{\textit{\textbf{Double Experts}}} \\
% \multicolumn{19}{c}{\textit{Full Attention}} \\
% \textit{Full Attention}\\
\rowcolor{gray!10} 
DINOv2+CLIP    &8.4B  & \underline{79.0} & \bf63.1 & 69.8 & 57.7 & \bf86.4 & \underline{67.0} & \bf60.5 & \bf66.9 & 32.0 & \bf 64.7 \\
$\quad\Delta$DINOv2         &  &\cellcolor{pos}4.1  &\cellcolor{pos}1.5  &\cellcolor{pos}3.7  &\cellcolor{pos}11.5  &\cellcolor{pos}1.8  &\cellcolor{pos}9.1  &\cellcolor{pos}11.8  &\cellcolor{pos}3.5   &\cellcolor{pos}8.6  &   &\\
$\quad\Delta$CLIP           &  &\cellcolor{pos}0.5  &\cellcolor{pos}1.1  &\cellcolor{pos}3.0  &\cellcolor{neg}0.5  &\cellcolor{pos}0.5  &\cellcolor{pos}4.0  &\cellcolor{pos}3.1  &\cellcolor{pos}0.7   &\cellcolor{pos}1.5  &   &\\
\midrule
\rowcolor{gray!10} 
LayoutLMv3+CLIP&7.7B  & \bf79.2 &\underline{62.4} & 68.5 & \bf58.9 & \underline{86.1} & \bf67.0 & \underline{59.9} & \underline{66.8} & \bf33.0 & \underline{64.6} \\
$\quad\Delta$LayoutLMv3     &  &\cellcolor{pos}34.3  &\cellcolor{pos}22.4  &\cellcolor{pos}5.7  &\cellcolor{pos}15.3  &\cellcolor{pos}27.0  &\cellcolor{pos}38.0  &\cellcolor{pos}40.1  &\cellcolor{pos}32.0  &\cellcolor{pos}21.2  &   &\\
$\quad\Delta$CLIP           &  &\cellcolor{pos}0.7  &\cellcolor{pos}0.4  &\cellcolor{pos}1.7  &\cellcolor{pos}0.7  &\cellcolor{pos}0.2  &\cellcolor{pos}4.0  &\cellcolor{pos}2.5  &\cellcolor{pos}0.6   &\cellcolor{pos}2.5  &   &\\
\midrule
\rowcolor{gray!10} 
ConvNeXt+CLIP  &7.5B  & 78.7 & 61.9 & \underline{69.9} & \underline{57.8} & 86.1 & 65.5 & 59.2 & 66.1 & \underline{32.1} &64.1 \\
$\quad\Delta$ConvNeXt       &  &\cellcolor{pos}3.6  &\cellcolor{pos}1.4  &\cellcolor{pos}4.9  &\cellcolor{pos}1.5  &\cellcolor{pos}0.5  &\cellcolor{pos}2.2  &\cellcolor{pos}4.2  &\cellcolor{pos}4.6   &\cellcolor{pos}6.1  &   &\\
$\quad\Delta$CLIP           &  &\cellcolor{pos}0.2  &\cellcolor{neg}0.1  &\cellcolor{pos}3.1  &\cellcolor{neg}0.4  &\cellcolor{pos}0.2  &\cellcolor{pos}2.5  &\cellcolor{pos}1.8  &\cellcolor{neg}0.1   &\cellcolor{pos}1.6  &   &\\
\midrule
\rowcolor{gray!10} 
SAM+CLIP       &7.9B  & 75.4 & 60.5 & \bf71.6 & 53.4 & 85.4 & 65.4 & 57.5 & 62.0 & 29.1 &62.3 \\
$\quad\Delta$SAM            &  &\cellcolor{pos}10.7  &\cellcolor{pos}4.7  &\cellcolor{pos}7.7  &\cellcolor{pos}9.3  &\cellcolor{pos}3.4  &\cellcolor{pos}21.7  &\cellcolor{pos}23.6  &\cellcolor{pos}10.1   &\cellcolor{pos}11.4  &   &\\
$\quad\Delta$CLIP           &  &\cellcolor{neg}3.1  &\cellcolor{neg}1.5  &\cellcolor{pos}4.8  &\cellcolor{neg}4.8  &\cellcolor{neg}0.5  &\cellcolor{pos}2.4  &\cellcolor{pos}0.1  &\cellcolor{neg}4.2   &\cellcolor{neg}1.4  &   &\\
\bottomrule
\end{tabular}
}
\caption{Performance comparison of different double-expert methods. 
The $\Delta$-marked rows are compared to the single-expert method. 
Where \colorbox{pos}{blue cells} indicate the \textbf{double-expert} model is better, and \colorbox{neg}{red cells} indicate the \textbf{single-expert} model is better.}
\label{tab:double-exp}
\end{table*}

%% file: table/three_experts.tex
\begin{table*}[ht]
% \fontsize{15}{20}\selectfont
\setlength{\tabcolsep}{3pt}
\centering
\resizebox{\textwidth}{!}{
\begin{tabular}{lrrrrrrrrrrrrrrr}
\toprule
Model & Param.  & VQA$^{\text{v2}}$ & GQA & SQA$^\text{I}$ & VQA$^\text{T}$ & POPE & MMB & MMB$^{\text{CN}}$ & SEED$^\text{I}$  & MM-Vet &Avg. \\
\midrule 
\rowcolor{gray!10} \multicolumn{12}{c}{\textit{\textbf{Triple Experts}}} \\
\rowcolor{gray!10} 
ConvNeXt+LayoutLMv3+CLIP &7.9B  & 78.5 & \underline{63.3}   &\bf70.2 & \underline{58.0} & \bf87.3  & \underline{66.8} & \underline{58.9}  & 66.0 & 32.2 &\underline{64.6} \\
$\quad\Delta$ConvNeXt+CLIP            &  &\cellcolor{neg}0.2  &\cellcolor{pos}1.4  &\cellcolor{pos}0.3  &\cellcolor{pos}0.2  &\cellcolor{pos}1.2   &\cellcolor{pos}1.3  &\cellcolor{neg}0.3  &\cellcolor{neg}0.1 &\cellcolor{pos}0.1   &\\
$\quad\Delta$LayoutLMv3+CLIP          &  &\cellcolor{neg}0.7  &\cellcolor{pos}0.9  &\cellcolor{pos}0.9  &\cellcolor{pos}1.7  &\cellcolor{pos}1.2   &\cellcolor{neg}0.2  &\cellcolor{neg}1.0   &\cellcolor{neg}0.8 &\cellcolor{neg}0.8  &\\
\midrule 
\rowcolor{gray!10} 
ConvNeXt+DINOv2+CLIP     &8.6B  &  \underline{78.6}    & 63.2    & \underline{69.2} & 57.8 & 86.5  & 66.6 & 58.9  & \underline{67.1} & \underline{32.9} &64.5 \\
$\quad\Delta$ConvNeXt+CLIP            &  &\cellcolor{neg}0.1  &\cellcolor{pos}1.3  &\cellcolor{neg}0.7 &\cellcolor{pos}0.0  &\cellcolor{pos}0.4  &\cellcolor{pos}1.1   &\cellcolor{neg}0.3  &\cellcolor{pos}1.0  &\cellcolor{pos}0.8   &\\
$\quad\Delta$DINOv2+CLIP              &  &\cellcolor{neg}0.4  &\cellcolor{pos}0.1  &\cellcolor{neg}0.6  &\cellcolor{pos}0.1  &\cellcolor{pos}0.1   &\cellcolor{neg}0.4  &\cellcolor{neg}1.6   &\cellcolor{pos}0.2 &\cellcolor{pos}\cellcolor{pos}0.9 &     \\ 
\midrule
\rowcolor{gray!10}
LayoutLMv3+DINOv2+CLIP   &8.8B  & \bf79.1 & \bf63.6 & 69.0 &\bf58.4 & \underline{86.5}  &\bf67.4 &\bf60.0  &\bf67.5 &\bf33.6 &\bf 65.0\\
$\quad\Delta$LayoutLMv3+CLIP          &  &\cellcolor{neg}0.1  &\cellcolor{pos}1.2  &\cellcolor{pos}0.5  &\cellcolor{neg}0.5  &\cellcolor{pos}0.4   &\cellcolor{pos}0.4  &\cellcolor{pos}0.1   &\cellcolor{pos}0.7 &\cellcolor{pos}0.6   &\\
$\quad\Delta$DINOv2+CLIP              &  &\cellcolor{pos}0.1  &\cellcolor{pos}0.5  &\cellcolor{neg}0.8  &\cellcolor{pos}0.7  &\cellcolor{pos}0.1   &\cellcolor{pos}0.4  &\cellcolor{neg}0.5   &\cellcolor{pos}0.6 &\cellcolor{pos}1.6   & \\
\bottomrule
\end{tabular}
}
\caption{Performance comparison of different triple-expert methods.
The $\Delta$-marked rows are compared to the double-expert method. 
Where \colorbox{pos}{blue cells} indicate the \textbf{triple-expert} model is better, and \colorbox{neg}{red cells} indicate the \textbf{double-expert} model is better.}
\label{tab:triple-exp}
\end{table*}

%% file: table/fusion_methods.tex
\begin{table*}[ht]
\setlength{\tabcolsep}{3pt}
\centering
\resizebox{\textwidth}{!}{
\begin{tabular}{lcccccccccccccc}
\toprule
Model & Param.  & VQA$^{\text{v2}}$ & GQA & SQA$^\text{I}$ & VQA$^\text{T}$ & POPE & MMB & MMB$^{\text{CN}}$ & SEED$^\text{I}$  & MM-Vet \\
\midrule 
\rowcolor{gray!10} \multicolumn{11}{c}{\textit{\textbf{Effect of Fusion Methods}}} \\
DINOv2+CLIP+MLP    &8.4B  & \bf79.0 & \bf63.1 & \bf69.8 & \bf 57.7 & \bf86.4 & \bf 67.0 & \bf60.5 & \bf66.9 & \bf32.0  \\
DINOv2+CLIP+Q-Former    &8.5B  & 60.4 & 50.9 & 66.7 & 45.1 & 45.2 & 52.7 & 44.8 & 51.8 & 20.5  \\
\midrule 
% LayoutLMv3+CLIP+MLP&  & \bf79.2 & \bf62.4 & \bf49.6 & 68.5 & \bf58.9 & \bf86.1 & \bf1426.5 & \bf67.0 & \bf59.9 & \bf66.8 &  & \bf33.0  \\
% LayoutLMv3+CLIP+Q-former&  & 68.5 & 54.7 & 47.7 & \bf69.8 & 48.2 & 81.8 & 1277.0 & 59.5 & 50.1 & 53.8 &  & 24.0  \\
% \midrule 
ConvNeXt+CLIP+MLP  &7.5B  & \bf78.7 & \bf61.9 &\bf 69.9 & \bf57.8 & \bf86.1 & \bf65.5 & \bf 59.2 & \bf66.1 & \bf32.1  \\
ConvNeXt+CLIP+Q-Former  &7.6B  & 65.8 & 52.6 & 68.7 & 45.6 & 77.0 & 59.7 & 49.8 & 53.2 & 22.1  \\
\bottomrule
\end{tabular}
}
\caption{Performance comparison of different poly-expert fusion methods.}
\label{tab:fusion-exp}
\end{table*}

%% file: table/experts_order.tex
\begin{table*}[ht]
\setlength{\tabcolsep}{3pt}
\centering
\resizebox{\textwidth}{!}{
\begin{tabular}{lcccccccccccccc}
\toprule
Model & Param.  & VQA$^{\text{v2}}$ & GQA & SQA$^\text{I}$ & VQA$^\text{T}$ & POPE & MMB & MMB$^{\text{CN}}$ & SEED$^\text{I}$  & MM-Vet \\
\midrule 
\rowcolor{gray!10} \multicolumn{11}{c}{\textit{\textbf{Effect of the Order of Experts}}} \\
DINOv2$\to$CLIP    &8.4B  & 79.0 & 63.1 & \bf69.8 & 57.7 & \bf86.4 & 67.0 & \bf60.5 & 66.9 & \bf 32.0  \\
CLIP$\to$DINOv2  &8.4B  & \bf79.6 & \bf63.9 & 69.2 & \bf59.1 & 86.4 & \bf67.5 & 59.4 & \bf 67.0 & 31.8 \\
\midrule 
ConvNeXt$\to$CLIP  &7.5B  & \bf78.7 &\bf 61.9 &\bf 69.9 &\bf 57.8 & 86.1 & 65.5 & \bf59.2 & \bf66.1 & \bf32.1  \\
CLIP$\to$ConvNeXt    &7.5B  & 78.0 & 61.9 & 68.7 & 57.4 & \bf86.9 &\bf 66.0 & 58.1 & 65.4 & 30.6  \\

\bottomrule
\end{tabular}
}
\caption{Performance comparison of different expert orders. We exchange the order of experts in ``DINOv2+CLIP'', and ``ConvNext+CLIP''.}
\label{tab:order-exp}
% \vspace{-15pt}
\end{table*}

%% file: table/PE-table.tex
\begin{table*}[ht]
\fontsize{15}{20}\selectfont
\setlength{\tabcolsep}{3pt}
\centering

\resizebox{\textwidth}{!}{
\begin{tabular}{lccccccccccccc}
\toprule

Model   & VQA$^{\text{v2}}$ & GQA  & SQA$^\text{I}$ & VQA$^\text{T}$ & POPE & MMB & MMB$^{\text{CN}}$ & SEED$^\text{I}$  & MM-Vet & Avg.\\
\midrule 
\rowcolor{gray!10} \multicolumn{11}{c}{\textit{\textbf{Different Positional Encoding Schemes}}} \\
Origin      & 78.5 & 62.0  & 66.8 & \underline{58.2} & 85.9  & 64.3 & \underline{58.3} & \underline{66.2}  &30.5  &63.4\\
Share-all     &\underline{79.0}  &\underline{62.4}  &\bf68.4 &\bf58.4  &\underline{86.3}  &\bf67.4  &58.2  &65.7   &\underline{31.7}  &\bf64.2\\
Share-by-row     &75.0  &57.2  &63.4  &51.7  &86.1  &46.4  &43.4  &55.6   &\bf31.9 &56.7 \\
Share-by-row\&col   &\bf79.0  &\bf62.6  &\underline{68.3}  &58.1  &\bf86.3  &\underline{66.0}  &\bf58.8  &\bf66.2   &30.6 &\underline{64.0} \\
\bottomrule
\end{tabular}
}
\caption{{\bf Comparison of four positional encoding schemes on 9 benchmarks.}}
\label{tab:PE-exp}
% \vspace{-15pt}
\end{table*}

%% file: table/attention.tex
\begin{table*}[t]
\centering
\begin{tabular}{lrrrr}
\toprule
Benchmark & Text Prompt  & LayoutLMv3 & DINOv2 & CLIP \\
\midrule 
MMB & 61.1\% & 0.14\% & 2.76\% & 11.1\% \\
MMB$^{\text{CN}}$ & 58.8\% & 0.16\% & 2.92\% & 10.7\% \\
\bottomrule
\end{tabular}
\caption{Average attention probability (\%) allocation of Mousi's output on each visual expert. The model used is a ``LayoutLMv3+DINOv2+CLIP'' triple-expert visual encoder.}
\label{tab:atten-exp}
\end{table*}

%% file: figure/qa_case.tex
\begin{figure*}[t!]
  \centering
  \includegraphics[width=\textwidth]{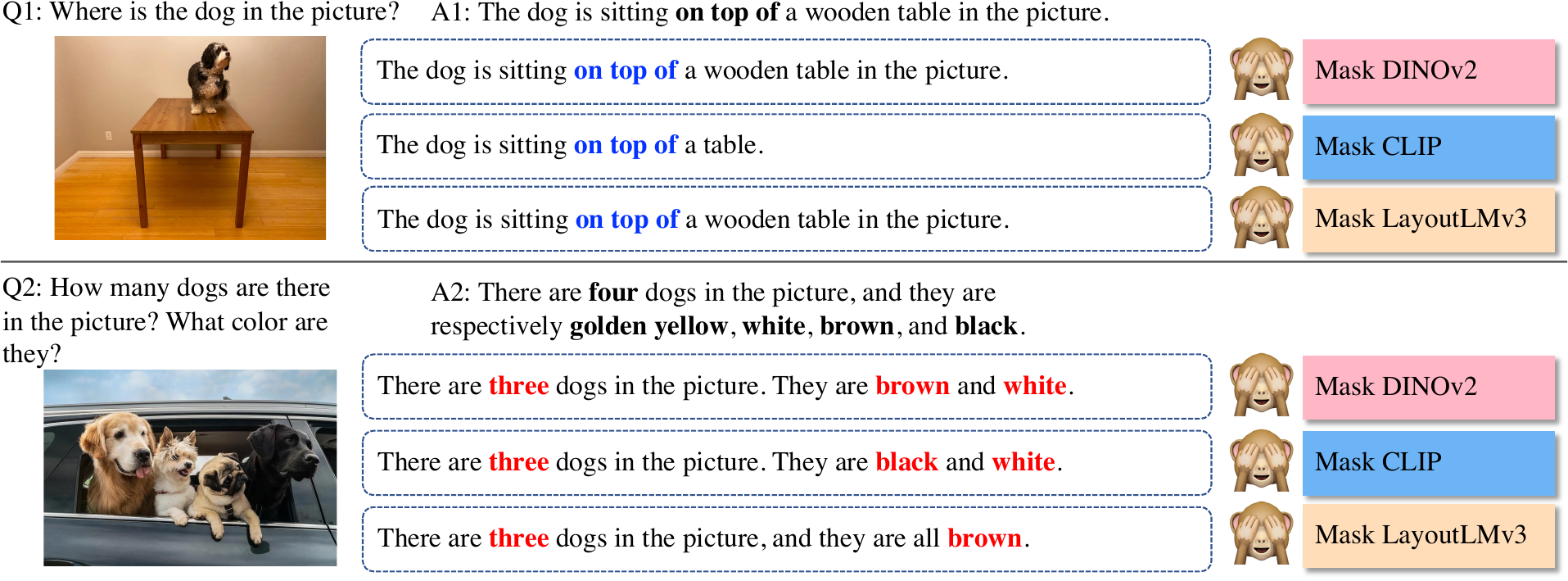}
  \caption{
  The perturbation experiments on the triple-expert LayoutLMv3+DINOv2+CLIP model, the specific perturbation is to mask all the output of the corresponding vision expert.
  }
  \label{fig:qa_case}
\end{figure*}

%% file: table/extend-data.tex
\begin{table*}[ht]
\fontsize{15}{20}\selectfont
\setlength{\tabcolsep}{3pt}
\centering

\resizebox{\textwidth}{!}{
\begin{tabular}{lcccccccccccccc}

\toprule
Model & Param.  & VQA$^{\text{v2}}$ & GQA  & SQA$^\text{I}$ & VQA$^\text{T}$ & POPE & MMB & MMB$^{\text{CN}}$ & SEED$^\text{I}$  & MM-Vet &Avg.\\
\midrule 
\rowcolor{gray!10} \multicolumn{12}{c}{\textit{\textbf{Baselines}}} \\
InstructBLIP\cite{dai2023instructblip} & 8.0B & -- & 49.2  & 60.5 & 50.1 & --  & 36.0 & 23.7 & 53.4  & 26.2 & \\
Qwen-VL-Chat\cite{bai2023qwen} & 9.6B  & 78.2 & 57.5 & 68.2 &\bf 61.5 & --  &60.6 &56.7  &58.2 & -- &\\
BLIP-2\cite{li2023blip} &14.1B   & 41.0 & 41.0 & 61.0 &42.5 &85.3  & -- &--  &46.4 & 22.4 &\\
Shikra\cite{chen2023shikra} &7.3B   & 77.4 & -- & -- & -- &--  & 58.8 &--  &-- & -- &\\
PandaGPT\cite{su2023pandagpt} &13B   & -- & -- & -- & -- &--  & 45.4 &32.0  &47.6 & 19.6 &\\
mPLUG-Owl2\cite{ye2023mplugowl2} &8.2B   & -- & -- & -- & -- &--  & 66.5 &59.5  &64.5 & \underline{35.7} &\\
Emu2-chat\cite{sun2023generative} &37B   & -- & -- & -- & -- &--  & 62.4 &44.2  &68.9 & 31.0 &\\
% LLaVA-1.5 &7.3B  & 78.5 & 63.6 & 69.0 &58.4 & 86.5  &67.4 &60.0  &67.5 &33.6 \\

\rowcolor{gray!10} \multicolumn{12}{c}{\textit{\textbf{Default Data}}} \\
CLIP (LLaVA-1.5\cite{liu2023improvedllava}) & 7.3B & 78.5 & 62.0  & 66.8 & 58.2 & 85.9  & 64.3 & 58.3 & 66.2  & 30.5 & 63.1 \\
% LayoutLMv3+DINOv2+CLIP   &8.8B  & 79.1 &\bf 63.6 & 69.0 &58.4 & \bf86.5  &67.4 &60.0  &67.5 &33.6 \\
ConvNeXt+LayoutLMv3+CLIP   &7.9B  & 78.5 &\bf 63.3 & 70.2 & 58.0 & \bf87.3  &66.8 &58.9  &66.0 &32.2 &64.6\\

\rowcolor{gray!10} \multicolumn{12}{c}{\textit{\textbf{Data Enhancement}}} \\
CLIP  & 7.3B & \underline{80.8} &  62.7 & \underline{81.9} & 60.7 & 85.5 & \bf 69.2 & \underline{61.7} & \underline{69.8} & 35.6 &\underline{67.5}  \\
LayoutLMv3+ConvNeXt+CLIP  & 7.9B & \bf 80.9 & \underline{62.6} & \bf 84.3 & \underline{61.3} &  \underline{86.3} & \underline{68.8} & \bf 63.7 & \bf 70.1 & \bf 38.4  & \bf68.5\\
\bottomrule
\end{tabular}
}
\caption{The effect of data enhancement on nine benchmarks. \textbf{Param.} indicates the number of parameters.}
\label{tab:enhance-exp}
\end{table*}

%% file: outline/case.tex
\section{Case Study}

\input{figure/case_study}

Figure~\ref{fig:case0} shows the case study of MouSi on seven tasks, including Complex Image Captioning, Visual Text Generating, OCR Interpreting Reasoning with World Knowledge, Visual Math Problem Solving, Complex Counting, and Visual Grounding. MouSi is able to successfully follow a variety of multimodal instructions, allowing for flexible interaction with humans.

%% file: figure/case_study.tex
\begin{figure*}[t!]
  \centering
  \includegraphics[width=\textwidth]{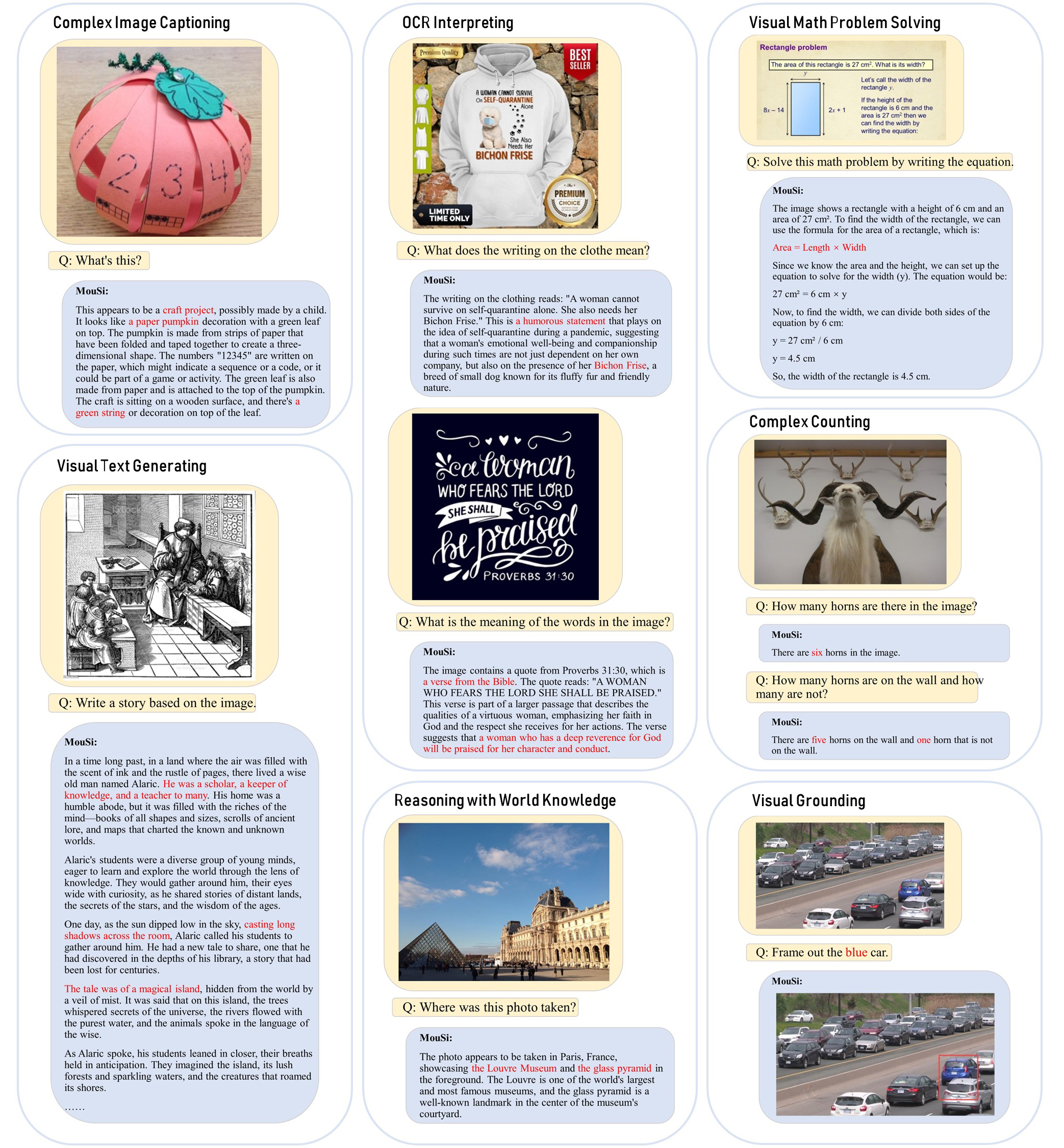}
  \caption{Qualitative examples generated by Mousi.
  }
  \label{fig:case0}
\end{figure*}

%% file: outline/related_work.tex
\section{Related Work}

\paragraph{Vision-Language Models (VLMs)} represent the confluence of linguistic and visual processing, and they have shown promising results in various applications. Early models such as VisualGPT \cite{chen2022visualgpt} provided foundational work in image captioning, while the BLIP series \cite{li2022blip,li2023blip} extended capabilities to include visual question answering. Flamingo \cite{alayrac2022flamingo} and Kosmos-1 \cite{huang2023language} demonstrated effective multi-modal understanding within image-text frameworks. LLaMA adaptations like LLaVA \cite{liu2023llava} and MiniGPT-4 \cite{zhu2023minigpt} utilize projection layers for connecting vision encoders and LLMs.
 CoGVLM \cite{wang2023cogvlm} replicated close to double the parameters to build visual experts specializing in visual tokens, while similar to our exploration of positional encoding, they used share-by-one rather than the original approach.
 Qwen-VL and BLIP series \cite{bai2023qwen,dai2305instructblip} use the Q-Former network to bridge text and image.
 
\paragraph{Visual Encoding Experts} 
The success of vision language models pivots upon adept visual encoding; hence, a curated selection of vision encoders, each with its own domain expertise, is crucial for holistic visual understanding.
The CLIP model by \cite{radford2021learning} employs contrastive learning to align images and text, effectively facilitating semantic image understanding.
Dinov2 \cite{oquab2023dinov2} from Meta advances self-supervised learning through a student-teacher network paradigm, developing spatial understanding with a robust ViT framework.
Microsoft's LayoutLMv3 \cite{huang2022layoutlmv3}, on the other hand, presents a multimodal Transformer adept in document AI by bolstering word-patch alignment in a ViT model.
Convnext \cite{woo2023convnext} reintroduces the efficacy of ConvNets with its FCMAE framework and GRN layer, finetuned with ImageNet-22K data.
The Segment Anything Model (SAM) by \cite{kirillov2023segment} showcases exceptional segmentation prowess, trained on a vast dataset to champion zero-shot generalization in its ViT infrastructure.
The MAE \cite{he2021masked} demonstrated remarkable denoising self-supervised capabilities, reconstructing images with a high degree of fidelity.
Yet these encoders, notably CLIP, possess limitations as evidenced by \cite{kamath2023s} highlighted its struggles with spatial orientation and \cite{li2023evaluating}'s findings on object hallucination. Moreover, \cite{wang2023makes} recognized a division of competencies, noting more semantics in fully/weakly supervised encoders like CLIP, while others excel in fine-grained perception. 

\paragraph{Multi-Modal Large Language Models (MLLMs)} have been evolving rapidly, with models like ImageBind-LLM \cite{han2023imagebind} and PandaGPT \cite{su2023pandagpt} incorporating richer modality inputs, including audio and video. There is also a growing focus on region-level parsing \cite{chen2023shikra}, text-to-image generation \cite{wen2023improving}, and 3D understanding \cite{xu2023pointllm}. These models show that MLLMs can achieve meaningful performance across a range of tasks. 
MouSi, as a poly-visual-expert VLM, is easily adapted to multi-modal-expert models, which will be our future work.

%% file: outline/appendix.tex
\section{Datasets}
\label{app:data}

During the pretrain phase, we employed the identical LCS-558K dataset as utilized in LLaVA-1.5, sourced from LAION-CC-SBU. For data-enhanced datasets, we incorporated the pre-trained dataset from ShareGPT4V~\cite{chen2023sharegpt4v}, distinguished by its longer textual descriptions.

In the subsequent finetune phase, we utilized the same instruction-based fine-tuning data as LLaVA-1.5 for the default dataset, comprising approximately 665K samples. For datasets with enhanced data, we introduced supplementary data during the finetune stage, drawing from sources such as ShareGPT4V, LVIS-INSTRUCT4V~\cite{wang2023see}, and CogVLIM-SFT-311K-CN~\cite{wang2023cogvlm}.

The specifics of our pretrain and finetune datasets are detailed in Table~\ref{tab:data_mixture}.

% Default instruction data, data enhanced instruction data
\begin{table}[h!]
\centering
\begin{tabular}{p{40mm} l| p{40mm} l}
\toprule
Default pretrain data & Size & Enhanced pretrain data & Size \\
\midrule
LCS-558K & 558K & ShareGPT4V & 1200K \\
\bottomrule
\\
\toprule
Default finetune data & Size & Enhanced finetune data & Size \\
\midrule
LLaVA & 158K & ShareGPT4V-cap100k & 100K \\
ShareGPT & 40K & ShareGPT4V-mix-665k & 665K \\
VQAv2 & 83K & LVIS-INSTRUCT4V-220k & 220K  \\
GQA & 72K & CogVLM-SFT-311K-CN & 150K \\
OCRVQA & 80K & VG & 86K \\
A-OKVQA & 50K & OCRVQA & 80K \\
TextCaps & 22K & GQA & 72K \\
RefCOCO & 30K & VQAv2 & 60K \\
VG & 86K &   docvqa & 44K \\
 OKVQA & 9K  & stvqa & 30K \\
  &  & fmiqa & 23K \\
 &  &   textvqa & 21K \\
 &  & coco-cn & 20K  \\
 &  &  ScienceQA & 10K  \\
 &  & flickr8k-cn & 8K  \\
 &  &  chinese-food & 1K  \\
 % &  &  web-celebrity & 0.9K  \\
 % &  & web-landmark & 0.5K  \\
 % &  &  share-textvqa & 0.5K  \\
\midrule
Total & 665K & Total & 1647K \\
\bottomrule
\end{tabular}
\vspace{2mm}
\caption{
Default data and Enhanced data for the Pretrain and Finetune phases of our 
model.
}
\label{tab:data_mixture}
\end{table}

\section{Hyperparameters}
\label{app:hyper}

We use the same set of hyperparameters as the original LLaVA-1.5. The training hyperparameters for visual language alignment pre-training and visual instruction tuning are shown in Table~\ref{tab:hyperparameter}.

\begin{table}[h!]
\centering
\scalebox{0.96}{
\begin{tabular}{l| c c}
\toprule
Hyperparameter & Pretrain & Finetune \\
\midrule
batch size & 256 & 128 \\
lr & 1e-3 & 2e-5 \\
lr schedule & \multicolumn{2}{c}{cosine decay} \\
lr warmup ratio & \multicolumn{2}{c}{0.03} \\
weight decay & \multicolumn{2}{c}{0} \\
epoch & \multicolumn{2}{c}{1} \\
optimizer & \multicolumn{2}{c}{AdamW} \\
DeepSpeed stage & 2 & 3 \\
\bottomrule
\end{tabular}
}
\vspace{2mm}
\caption{
Hyperparameters of our model's pretrain and finetune.
}
\label{tab:hyperparameter}
\end{table}

\section{More Case Studies}
\input{figure/case}

%% file: figure/case.tex
% \begin{figure*}[t!]
%   \centering
%   \includegraphics[width=\textwidth]{figure/case_study.jpg}
%   \caption{Qualitative examples generated by Mousi.
%   }
%   \label{fig:case0}
% \end{figure*}

\begin{figure*}[h]
  \centering
  \includegraphics[width=0.9\textwidth]{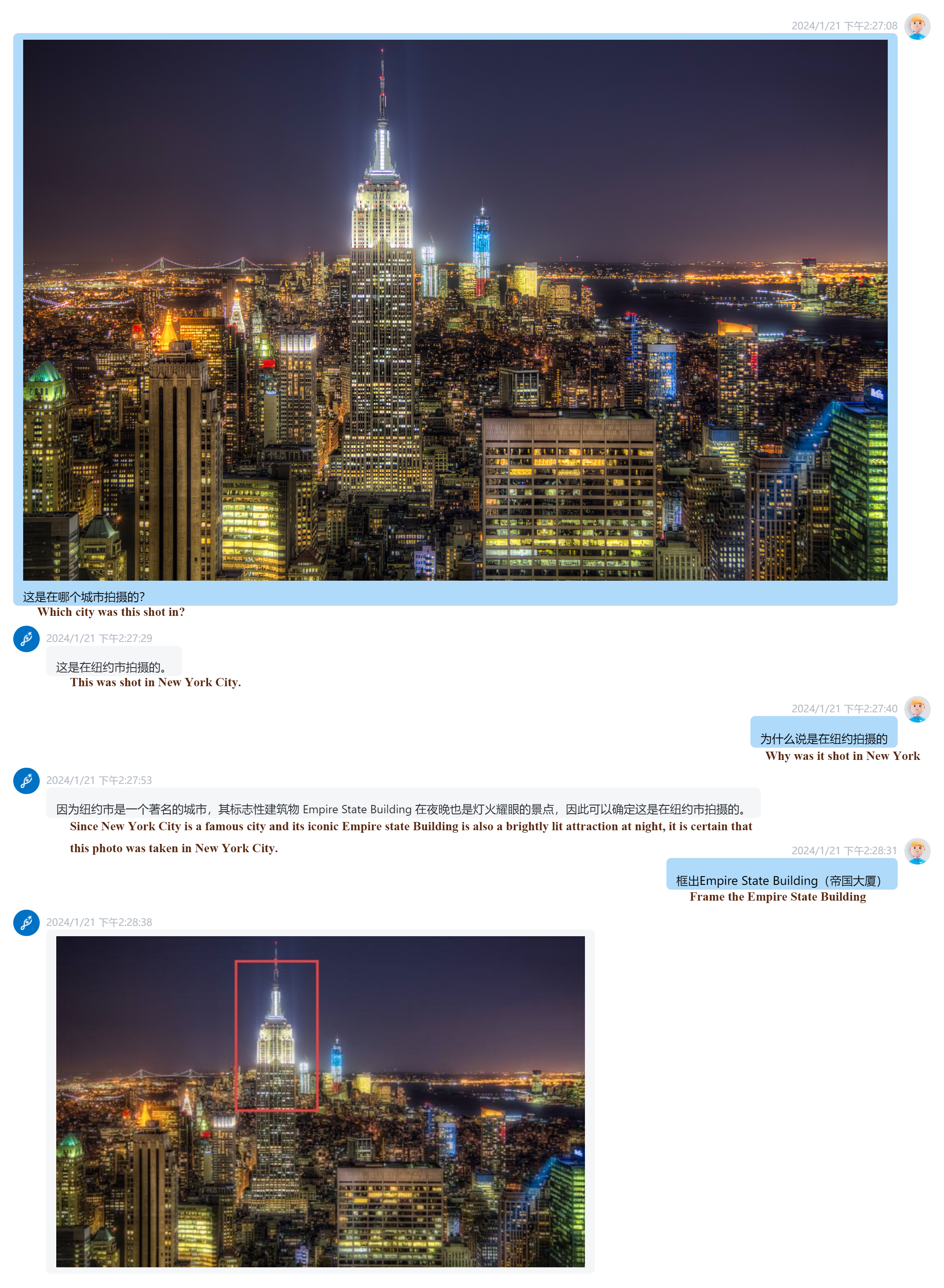}
  \caption{Qualitative Chinese examples generated by Mousi.
  }
  \label{fig:case1}
\end{figure*}
 
\begin{figure*}[t]
  \centering
  \includegraphics[width=0.9\textwidth]{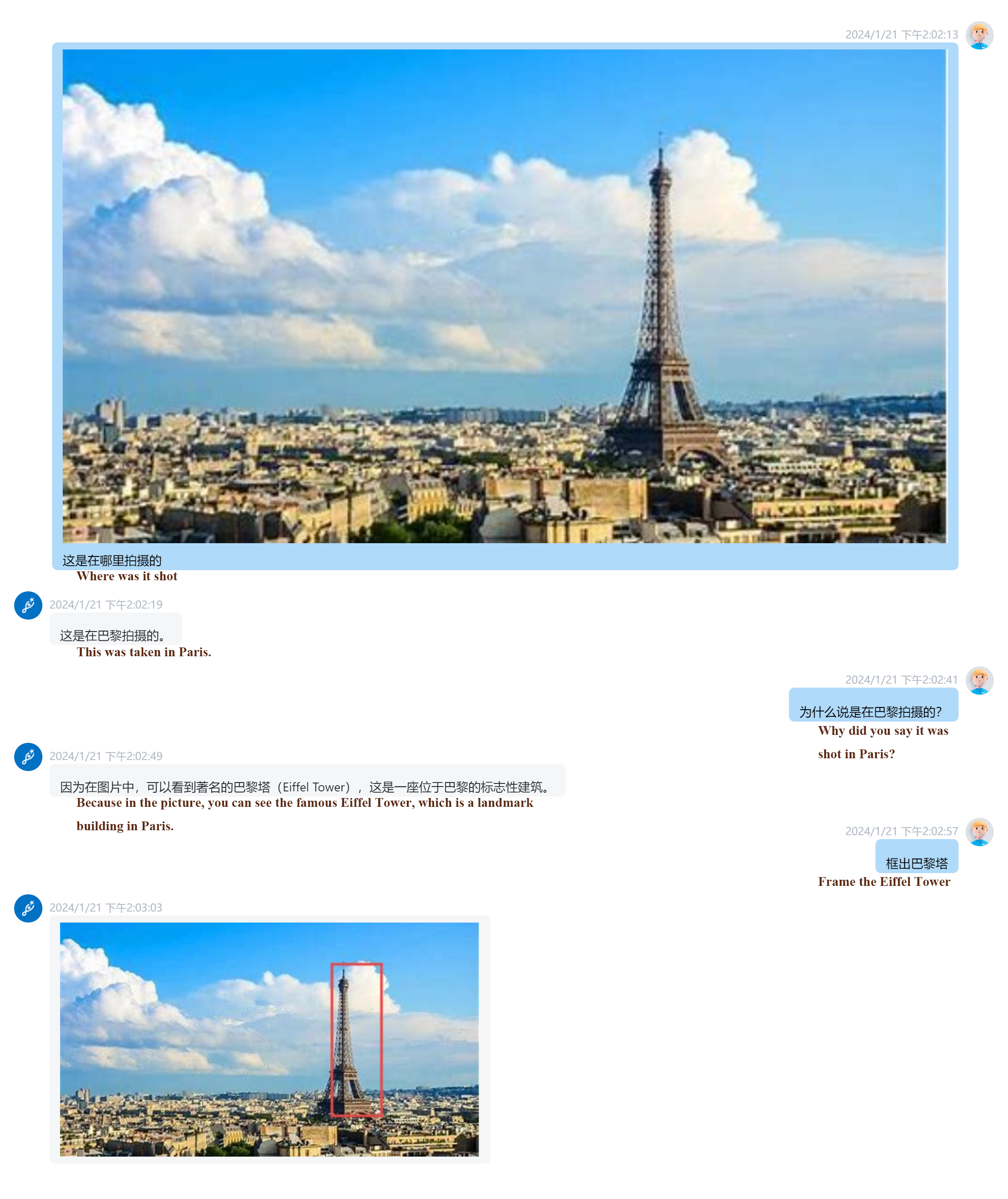}
  \caption{Qualitative Chinese examples generated by Mousi.
  }
  \label{fig:case2}
\end{figure*}
 
\begin{figure*}[t]
  \centering
  \includegraphics[width=0.9\textwidth]{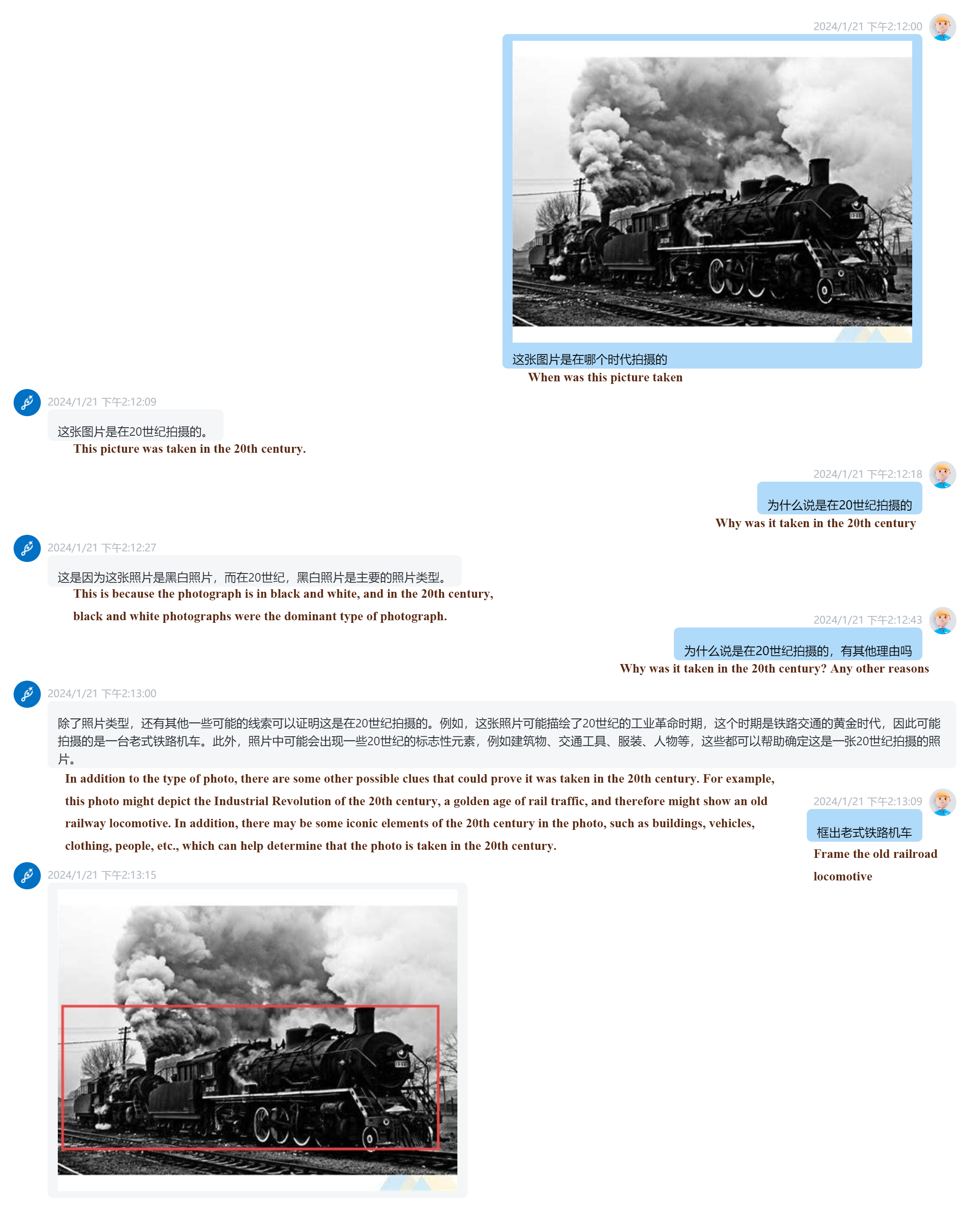}
  \caption{Qualitative Chinese examples generated by Mousi.
  }
  \label{fig:case3}
\end{figure*}

 \begin{figure*}[t]
  \centering
  \includegraphics[width=0.9\textwidth]{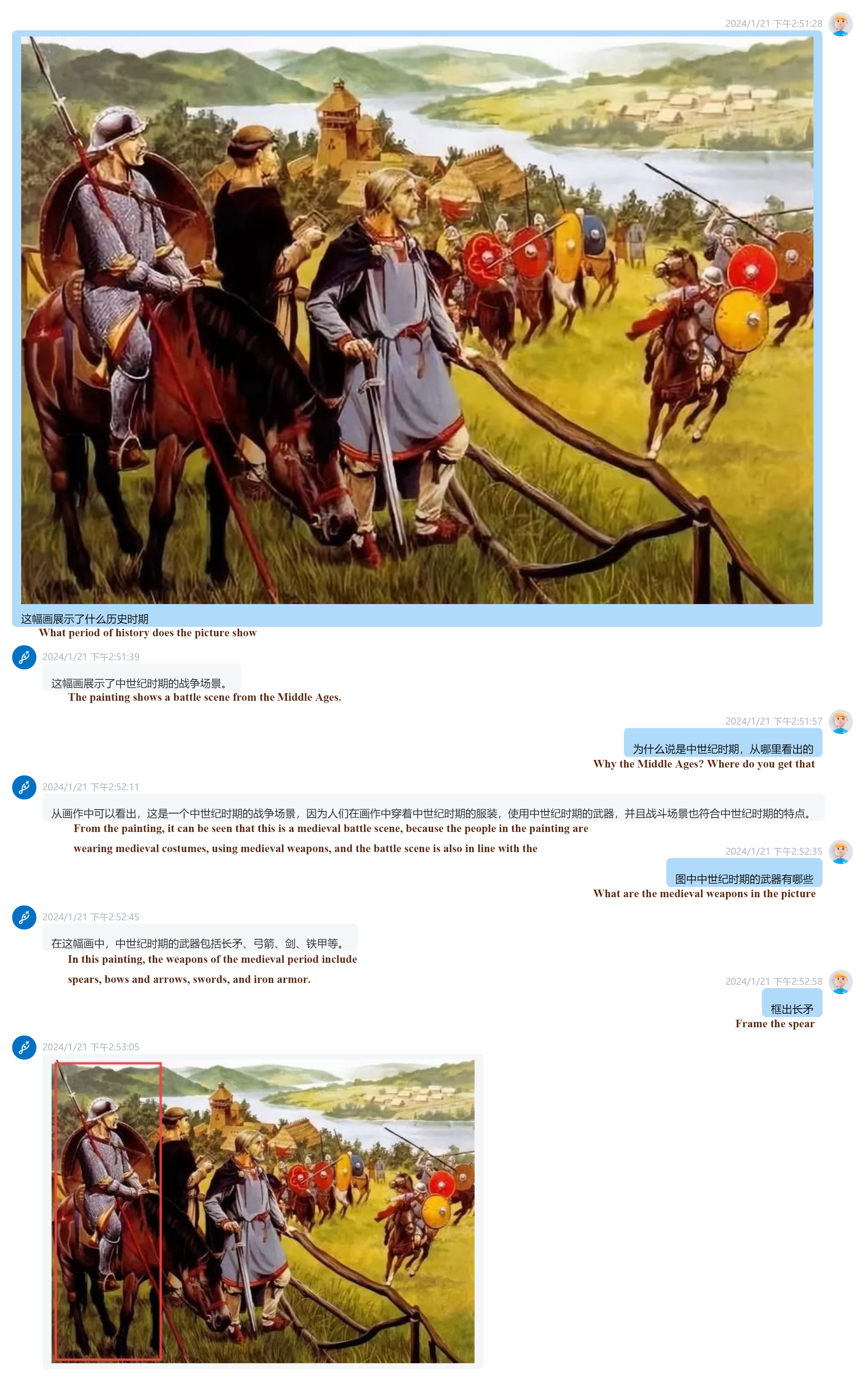}
  \caption{Qualitative Chinese examples generated by Mousi.
  }
  \label{fig:case4}
\end{figure*}

 \begin{figure*}[t]
  \centering
  \includegraphics[width=0.9\textwidth]{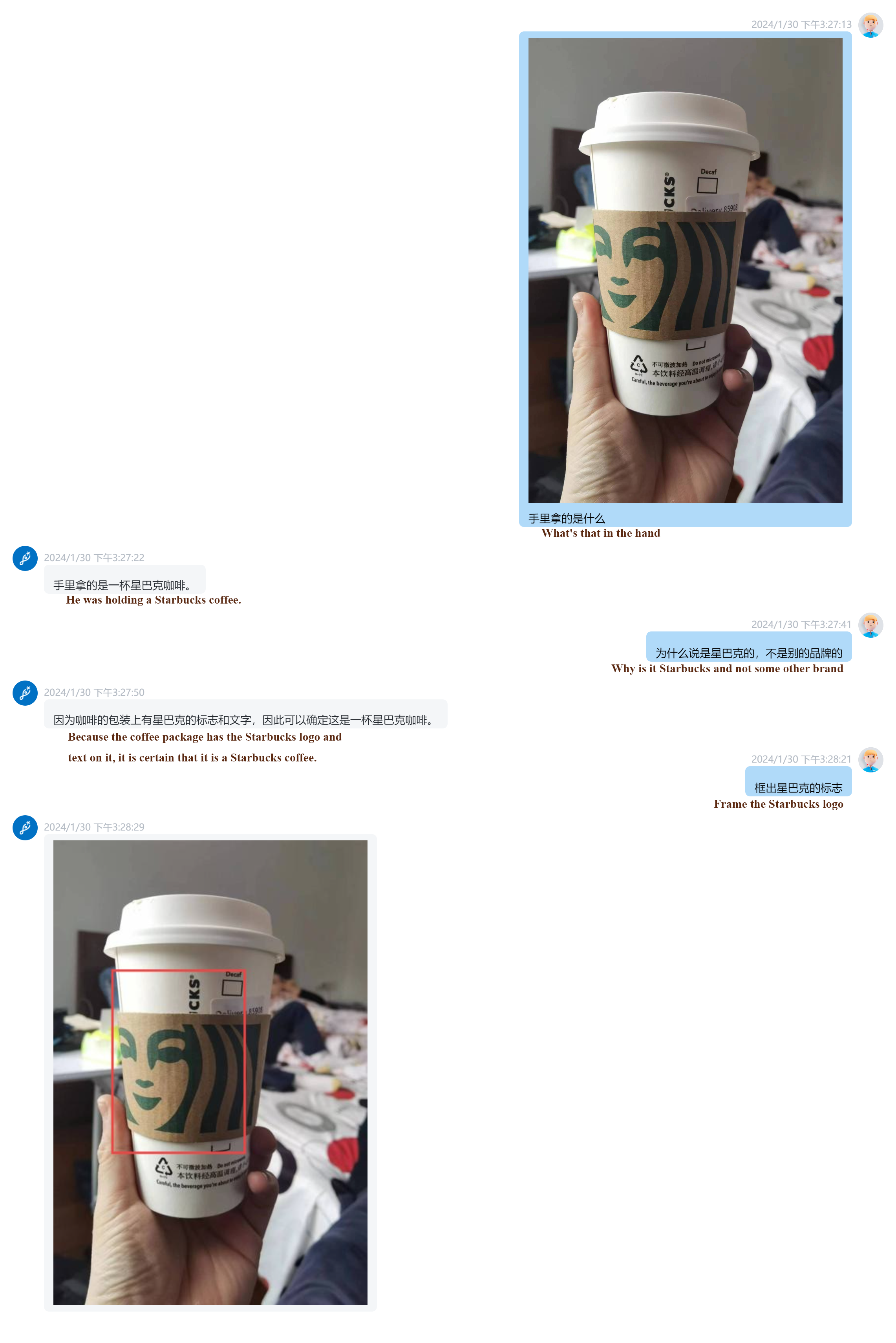}
  \caption{Qualitative Chinese examples generated by Mousi.
  }
  \label{fig:case5}
\end{figure*}

 \begin{figure*}[t]
  \centering
  \includegraphics[width=0.9\textwidth]{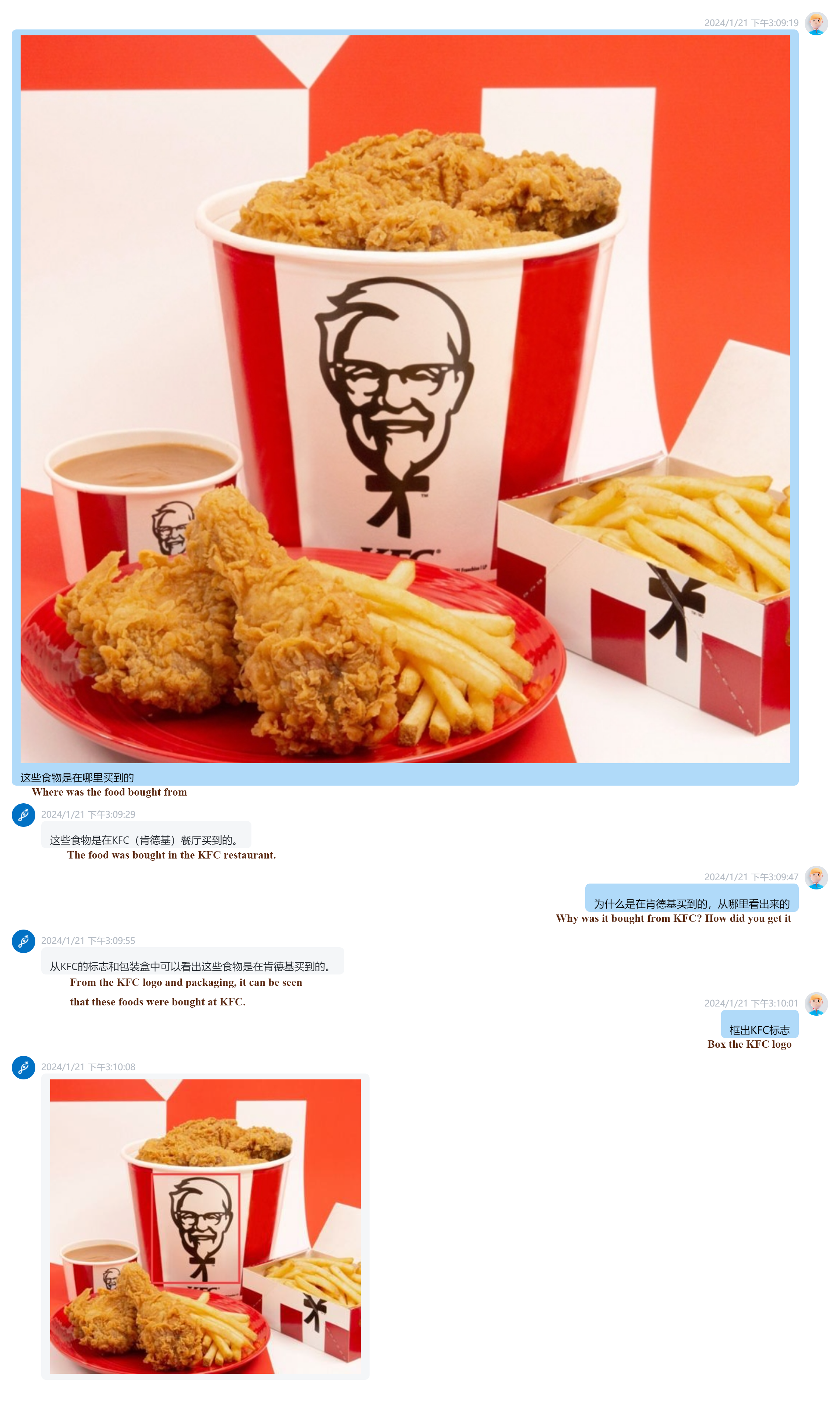}
  \caption{Qualitative Chinese examples generated by Mousi.
  }
  \label{fig:case6}
\end{figure*}